\documentclass[lettersize,journal]{IEEEtran}
\usepackage{algorithm}
\usepackage{array}
\usepackage{cite}
\usepackage{amsmath,amssymb,amsfonts}
\usepackage{textcomp}
\usepackage{xcolor}
\usepackage{booktabs}
\usepackage{subfloat}
\usepackage{graphicx}
\usepackage{amsthm}
\usepackage{float}
\usepackage{hyperref}

\newtheorem{theorem}{Theorem}
\newtheorem{proposition}{Proposition}
\usepackage{colortbl}
\usepackage{marvosym}
\usepackage{orcidlink}
\usepackage{amsthm}

\newtheorem{assumption}{Assumption}
\newtheorem{lemma}{Lemma}
\usepackage{algorithm}
\usepackage{algpseudocode}
\newtheorem{definition}{Definition}
\usepackage{adjustbox}
\usepackage{svg}
\usepackage{url}
\usepackage{verbatim}

\usepackage{cite}
\hypersetup{hidelinks}

\begin{document}

\title{Delayed Bottlenecking: Alleviating Forgetting in Pre-trained Graph Neural Networks}

\author{Zhe~Zhao\orcidlink{0000-0002-8942-8761},
  Pengkun~Wang\orcidlink{0000-0002-2680-4563}\textsuperscript{\Letter},~\IEEEmembership{Member,~IEEE},
  Xu~Wang\orcidlink{0000-0002-1492-3477},
  Haibin~Wen\orcidlink{0009-0009-5019-2390},
  Xiaolong~Xie\orcidlink{0000-0002-9016-6287},
  Zhengyang~Zhou\orcidlink{0000-0003-4728-7347},~\IEEEmembership{Member,~IEEE},
  Qingfu~Zhang\orcidlink{0000-0003-0786-0671},~\IEEEmembership{Fellow,~IEEE}
  and Yang~Wang\orcidlink{0000-0002-6079-7053}\textsuperscript{\Letter},~\IEEEmembership{Senior Member,~IEEE}

  \thanks{ \emph{Corresponding author: Prof. Yang Wang and Prof. Pengkun Wang.}}
  \thanks{Zhe Zhao, Pengkun Wang, Xu Wang, Zhengyang Zhou, and Yang Wang are with University of Science and Technology of China (USTC), Heifei 230022, China (e-mail: zz4543@mail.ustc.edu.cn; \{pengkun, wx309, zzy0929, angyan\}@ustc.edu.cn).}
  \thanks{Haibin Wen is with Shaoguan University, Shaoguan, China (haibin65535@gmail.com).}
  \thanks{Xiaolong Xie is with Nanchang University, Nanchang, China (416100210092@email.ncu.edu.cn).}
  \thanks{Qingfu Zhang is with the Department of Computer Science, City University of Hong Kong, Hong Kong, and also with the Shenzhen Research
Institute, City University of Hong Kong, Shenzhen 518057, China (e-mail: qingfu.zhang@cityu.edu.hk).}

}

\markboth{IEEE Transactions on Knowledge and Data Engineering,~Vol.~XX, No.~XX, XX~2024}%
{Zhao Zhe \MakeLowercase{\textit{et al.}}: Delayed Bottlenecking: Alleviating Forgetting in Pre-trained Graph Neural Networks}



\maketitle

\begin{abstract}
Pre-training GNNs to extract transferable knowledge and apply it to downstream tasks has become the de facto standard of graph representation learning. Recent works focused on designing self-supervised pre-training tasks to extract useful and universal transferable knowledge from large-scale unlabeled data. However, they have to face an inevitable question: traditional pre-training strategies that aim at extracting useful information about pre-training tasks, may not extract all useful information about the downstream task. In this paper, we reexamine the pre-training process within traditional pre-training and fine-tuning frameworks from the perspective of Information Bottleneck (IB) and confirm that the forgetting phenomenon in pre-training phase may cause detrimental effects on downstream tasks. Therefore, we propose a novel \underline{D}elayed \underline{B}ottlenecking \underline{P}re-training (DBP) framework which maintains as much as possible mutual information between latent representations and training data during pre-training phase by suppressing the compression operation and delays the compression operation to fine-tuning phase to make sure the compression can be guided with labeled fine-tuning data and downstream tasks. To achieve this, we design two information control objectives that can be directly optimized and further integrate them into the actual model design. Extensive experiments on both chemistry and biology domains demonstrate the effectiveness of DBP. The code is available in \url{https://anonymous.4open.science/r/TKDE-DBP}.
\end{abstract}


\section{Introduction}
\IEEEPARstart{I}{n} recent years, Graph Neural Networks (GNNs) have shown prominent performances in various fields including social networking~\cite{bhagat2011node,he2020bi,wang2021tedic,zhang2016understanding,zhang2020gcn}, molecular computing~\cite{gilmer2017neural,liu2019n,wu2023chemistry,reiser2022graph,li2022graph}, web recommendation~\cite{wang2019knowledge,ying2018graph,yang2023dgrec,chang2021sequential,reiser2022graph,hao2023multi,gao2022graph}, and  bioinformatics~\cite{shi2020graph,shi2019graphaf,you2018graph}.
Meanwhile, pre-training GNN, which is capable of enhancing the performance of GNN on specific-data-required downstream tasks by extracting universal transferable knowledge from large-scale unlabeled graph-structured data, has also attracted the great attention of both academic and industrial communities~\cite {ji2021survey,wang2023joint,zhao2024twist}.

Great efforts~\cite{hu2020gpt,DBLP:conf/kdd/HouLCDYW022,hu2020strategies,you2020graph,xu2021self,sun2020infograph,you2021graph} have been studied in the field of pre-training GNN to achieve knowledge extraction, and existing works can be roughly distinguished into two categories, contrastive self-supervised learning~\cite{you2020graph,xu2021self,sun2020infograph,you2021graph} and generative self-supervised learning~\cite{hu2020gpt,DBLP:conf/kdd/HouLCDYW022,hu2020strategies}. The previous one aims at learning knowledge in different semantic levels by contrasting the enhanced views of different data, while the latter one tries to recover and generate graph structure data to eventually learn the property patterns of vertexes and edges within the graph structure~\cite {wu2021self,tran2022s5cl}. In summary, all these methods have paid all their attention to the design of the self-supervised pre-training task to extract useful information with regard to the pre-training task from large-scale unlabeled datadata~\cite {zhu2021graph,verma2021graphmix,liu2023graphprompt,you2020graph,li2021targeted,wan2022factpegasus}. However, considering the difference between the pre-training task and downstream tasks, we have to face an inevitable question: \textit{can the pre-training process transfer all useful information to the downstream task from large-scale unlabeled data?}

\begin{figure}[!]
  \centering
   \includegraphics[width=\linewidth]{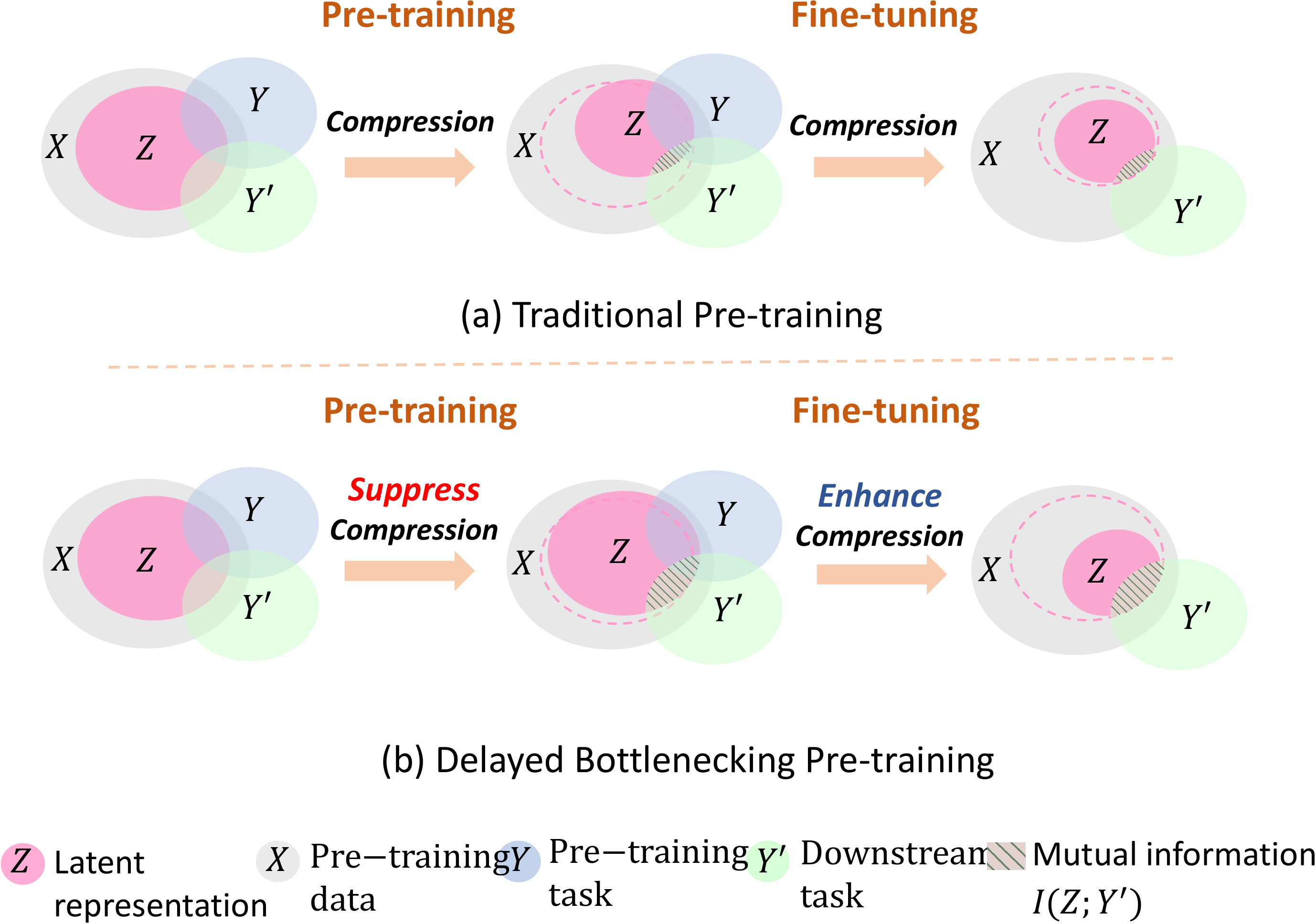}
   \caption{Information-theoretic analysis of conventional and delayed bottlenecking pre-training in graph neural networks. Subfigure (a) presents the dynamics of information encoding in latent space during conventional pre-training, denoted as \( Z \), relative to the pre-training data \( X \) and associated task \( Y \), and its subsequent impact on downstream task \( Y' \). In this regime, the latent representation \( Z \) undergoes a compression process, optimized for \( Y \), which inadvertently discards non-salient features for \( Y \) but may be pertinent to \( Y' \), thereby diminishing the mutual information \( I(Z; Y') \) post-compression. Subfigure (b) depicts an alternative approach with the proposed Delayed Bottlenecking Pre-Training, where the compression of \( Z \) during the pre-training phase is deliberately modulated. This control preserves a broader set of features in \( Z \), allowing for enhanced mutual information \( I(Z; Y') \) post-fine-tuning, which is refined under the guidance of labeled data specific to \( Y' \).}
   \label{fig:onecol}
\end{figure}

To answer this question, we need to re-examine the pre-training process within the traditional pre-training and fine-tuning framework from the perspective of information extraction. Some previous work indicates that, given a specific learning task, animals choose to forget some remembered behaviors to better adapt to some specific tasks~\cite{anderson2022prefrontal,gruber2019cultural,kitazono2017multiple,gravitz2019importance}. Meanwhile, ~\cite{achille2018critical,shwartz2017opening,li2020enhanced,peng2021learning} also indicate that this kind of biological forgetting phenomenon can also be found during the training process of neural networks.
As illustrated in Figure \ref{fig:onecol}(a), the neural network quickly learns information from data during the first phase and compresses the representation by forgetting some learned information which is useless to the pre-training task in the second phase. According to~\cite{shwartz2017opening}, such forgetting behavior is to better fit the target of the pre-training task. Nevertheless, considering the pre-training task which is artificially designed to extract universal transferable information from unlabeled data and is totally different from the downstream task in general~\cite{cossu2022continual,feng2020codebert}, such forgetting behavior is harmful to the learning and transformation of universal knowledge since those dropped information may be useful and of significance to downstream tasks. 

\textbf{Challenges.} To solve the above-mentioned deficiencies of traditional pre-training GNNs, there is a key challenge needs to be addressed: \textit{How to improve existing pre-training and fine-tuning strategy to make sure that the useful information with regard to a downstream task can be maintained as much as possible?} 

To address these challenges, as demonstrated in Figure \ref{fig:onecol}(b), we propose a novel \textbf{D}elayed \textbf{B}ottlenecking \textbf{P}re-training (DBP) framework to address the issue of information forgetting during pre-training. In particular, as illustrated in Figure \ref{fig:onecol}(b), we first re-analyze the whole procedure of pre-training from the perspective of IB, and formulate the information dropping during pre-training. Based on this,  we first design a novel information compression delayed pre-training strategy that maintains as much as possible mutual information between latent representations and training data during the pre-training phase by suppressing the compression operation. Then, we delay the compression operation to fine-tuning phase to make sure the compression can be guided with labeled fine-tuning data and downstream tasks. In the pre-training phase, DBP includes a newly designed information-based representation reconstruction which can maintain the mutual information between latent representation and training data by decoding the learned latent representation into the features of vertexes and edges. In the fine-tuning phase, we borrow the core idea of Depth Variational Information Bottleneck (DVIB)~\cite{shwartz2017opening} and extend it to be adapted to graph-structured data, hence making sure that the delayed information compression is optimized with the guidance of labeled fine-tuning data and downstream task. Extensive experiments on both chemistry and biology domains verify the effectiveness of our proposed strategy on various pre-training GNNs.

The main contributions are summarized as follows:
\begin{itemize}
    \item \textit{\textbf{New theoretical analysis:}} For the first time, we analyze the information forgetting of the compression operation in pre-training GNNs from the perspective of the Information Bottleneck (IB) theory. To our knowledge, this is the first paper that aims at alleviating the influence of such kind of inevitable information forgetting on downstream tasks.
    
    \item \textit{\textbf{Novel framework and methods:}} We propose a DBP framework that includes a novel information compression delayed pre-training strategy to enhance the performances of pre-training GNNs. In DBP, we propose a novel information-based representation reconstruction and an extended Graph-DVIB to respectively achieve information maintaining in the pre-training phase, and labeled fine-tuning data and downstream task-guided information compression in the fine-tuning phase.
    
    \item \textit{\textbf{Extensive empirical evaluation:}} Extensive experiments on both chemistry and biology domains demonstrate the effectiveness of our proposed framework while incorporating different pre-training GNNs.
\end{itemize}
    
    

\section{Related Work}
\noindent\textbf{Pre-training GNNs.}
Recently, Pre-training GNNs have received significant attention since they can alleviate the heavy reliance of traditional GNNs on data with fine-grained labels. Generally, pre-training GNNs usually consist of two phases: i) \textit{Pre-training}: learning model parameters and node embeddings from large-scale unlabeled graph data; ii) \textit{Fine-tuning}: fine-tuning the learned parameters and embeddings with labeled graph data to make the network more applicable to downstream tasks. 
Existing methods, which mostly follow such a two-phase framework, can be roughly divided into two categories: contrastive pre-training~\cite{you2020graph,xu2021self,velickovic2019deep} and generative pre-training~\cite{hu2020gpt,DBLP:conf/kdd/HouLCDYW022}. Contrastive pre-training learns graph representation by contrasting the semantic differences between pre-define positive and negative samples. In particular,  DGI~\cite{velickovic2019deep} focuses on the correspondences between nodes and subgraphs, GraphLoG~\cite {xu2021self} pays attention to correlogram and subgraph pairs, and GraphCL~\cite{xu2021self} contrasts subgraph level with different data augmentations.
On the other hand, generative pre-training captures the intrinsic dependencies between node attributes and graph structure by generating node attributes and edges, e.g., jointly generating nodes and edges~\cite{hu2020gpt} or reconstructing shadowed nodes~\cite{DBLP:conf/kdd/HouLCDYW022}. 
However, all these methods paid all their attention to designing self-supervised tasks to maximally extract useful information with regard to the pre-training task from large-scale unlabeled data, ignoring the issue that the knowledge extracted by the pre-training task cannot be completely transferred to the downstream task due to the differences between pre-training and downstream tasks. 

\noindent\textbf{Mutual Information and Its Application.}
Mutual information (MI) is a measure of the degree of interdependence between random variables based on Shannon entropy. It is often used to measure the nonlinear correlation between variables, so it can be regarded as a measure of the true dependence between variables. For two random variables $X$ and $Y$, the mutual information between them is as follows:
\begin{equation}
    \begin{aligned}
       I(X; Y) & = H(X) - H(X|Y) \\
    \end{aligned}
\end{equation}
Here, $H(X)$ is the information entropy of $X$, and $H(X|Y)$ is the conditional entropy of random variable $X$ under the condition of known random variable $Y$. From a probabilistic perspective, mutual information is derived from the joint probability distribution $p(x,y)$ and the marginal probability distribution $p(x)$ and $p(y)$ of the random variables $X$ and $Y$. The dependence between $X$ and $Y$ is stronger when the divergence between the joint probability distribution $p(x,y)$ and the marginal product $p(x)p(y)$ is larger. It is widely used in deep learning because it can measure the real dependencies between variables. 

However, since the true distribution in neural networks is difficult to know, the calculation and optimization of mutual information is a difficult problem. In graph learning, many studies on optimizing neural networks through information theory choose MINE~\cite{velickovic2019deep} or variational methods~\cite{shwartz2017opening} to approximate the upper and lower bounds of mutual information to achieve the goal of optimizing mutual information.
Many self-supervised learning methods on graphs also utilize mutual information. For example, DGI~\cite{velickovic2019deep} relies on maximizing mutual information between patch representations and corresponding high-level graph summaries to learn node representations in graph-structured data. GraphMVP~\cite{liu2021pre} performs self-supervised learning by optimizing the mutual information between molecular 2D topology and 3D geometric views to improve correspondence and consistency between these views. Besides representation learning and self-supervised learning, mutual information has also been used in the study of neural network interpretability and training dynamics. Typically, \cite{shwartz2017opening} investigates the correlation between changes in mutual information between representation and training data and labels during neural network training and the generalization and robustness of neural networks. These studies inspire us whether there will be related problems in the pre-training process of GNN.

\noindent\textbf{IB Theory.}
IB theory can be used in deep learning to seek the balance between fitting and generalization by controlling the mutual information between latent representation and training data. The main idea of such equilibrium can be summarized into two points: i) enlarging the information that is useful to the task within the representation, and ii) suppressing the information that is irrelevant to the task within the representation~\cite{shwartz2017opening,DBLP:conf/iclr/AlemiFD017,wu2020graph}. 
Given the significant potential of IB in enhancing model interpretability and generalization, recent researchers attempt to explore its effect on extracting graph representations~\cite{wu2020graph,yu2020graph,peng2020graph}. Specifically, GIB~\cite{wu2020graph} extends general IB to graph data as a modified regularization on both structure and feature information, hence achieving more robust node representations. And, SIB~\cite{yu2020graph} and VIB-GSL~\cite{peng2020graph} apply IB to subgraph recognition and graph structure learning, respectively. 
Collectively, these methods directly utilize the IB principle to learn minimal but sufficient information. However, directly using IB in pre-training GNNs to learn minimal but sufficient information with regard to the pre-training task will definitely result in the issue of information forgetting during the procedure of seeking the minimal information subset.

\section{Delayed Bottlenecking Pre-Training: Causation, Strategy, and Derivation}
A key insight of this paper is \textit{information suppressing in pre-training may lose useful information with regard to the downstream task.} In this section, we first conduct a theoretical analysis to demonstrate the existence of this effect and formulate the information forgetting during pre-training, and further pointedly propose the improvement strategy.
\subsection{Re-analyzing Parameter Transfer in Pre-training}

In this subsection, we re-analyze the information forgetting problem during pre-training and its impact on parameter transfer. Ideally, the essence of pre-training is to extract transferable knowledge from pre-training data, so the objective optimization process of pre-training can be described as:

\begin{equation}
    \begin{aligned}
    \theta_0 &= \mathop{\arg\min}\limits_{\theta}\;\mathcal{L}_{p}(f_\theta;\mathcal{D}^{pre}) 
    \\
    &= \mathop{\arg\max}\limits_{\theta}\;I_\theta(f_\theta;\mathcal{D}^{pre}),
    \end{aligned}
\end{equation}
where $\mathcal{L}_{p}$ represents the optimization objective of pre-training, and $I_\theta(f_\theta;\mathcal{D}^{pre})$ denotes the information extracted by model $f_\theta$ from pre-training dataset $\mathcal{D}^{pre}$. After training, the optimal model parameter set $\theta_0$ and the corresponding extracted information $I_\theta(f_\theta;\mathcal{D}^{pre})$ are transferred to the downstream task through parameter initialization.
\begin{lemma}[Representation Forgetting] \label{lemma:forgetting}
According to the research of ~\cite{shwartz2017opening}, in the normal training process, the mutual information $I(X;Z)$ between input data $X$ and latent representation $Z$ first increases and then decreases in the early stage of training, while the mutual information $I(X;Y)$ between input data $X$ and output $Y$ keeps increasing. Formally:
\begin{equation}
\begin{split}
\exists \theta_{max}, \text{ s.t. } \forall \theta < \theta_{max}, \frac{\partial I(X;Z_\theta)}{\partial \theta} > 0; \\
\forall \theta > \theta_{max}, \frac{\partial I(X;Z_\theta)}{\partial \theta} < 0.
\end{split}
\end{equation}
\end{lemma}

\begin{theorem}[Pre-training Information Transfer] \label{thm:info_tran}
Since the pre-training task itself is also a training task, according to Lemma \ref{lemma:forgetting}, for the existing pre-trained GNN model $f_\theta$, the extracted information $I_\theta(f_\theta;\mathcal{D}^{pre})$ will gradually increase to $I_{\theta_{max}}(f_{\theta_{max}};\mathcal{D}^{pre})$, and then decrease to $I_{\theta_{0}}(f_{\theta_{0}};\mathcal{D}^{pre})$ when obtaining the optimal parameter $\theta_0$, i.e.,
\begin{equation}
    \begin{aligned}
        &\quad\quad\quad\quad\quad\quad\quad\quad \theta \to \theta_0 \Longrightarrow 
        \\
        &I_\theta(f_\theta;\mathcal{D}^{pre}) \to I_{\theta_{max}}(f_{\theta_{max}};\mathcal{D}^{pre}) \to I_{\theta_{0}}(f_{\theta_{0}};\mathcal{D}^{pre}).
    \end{aligned}
\end{equation}
\end{theorem}
In this process, the forgotten information, probably contains some information which is useless to the pre-training task but useful to the downstream task. If such information is forgotten and cannot be transferred to the downstream task, the parameter set $\theta_0$ and extracted information $I_\theta(f_\theta;\mathcal{D}^{pre})$ are not optimal anymore.

\begin{figure*}[h]
    \centering
    \includegraphics[width=1 \linewidth]{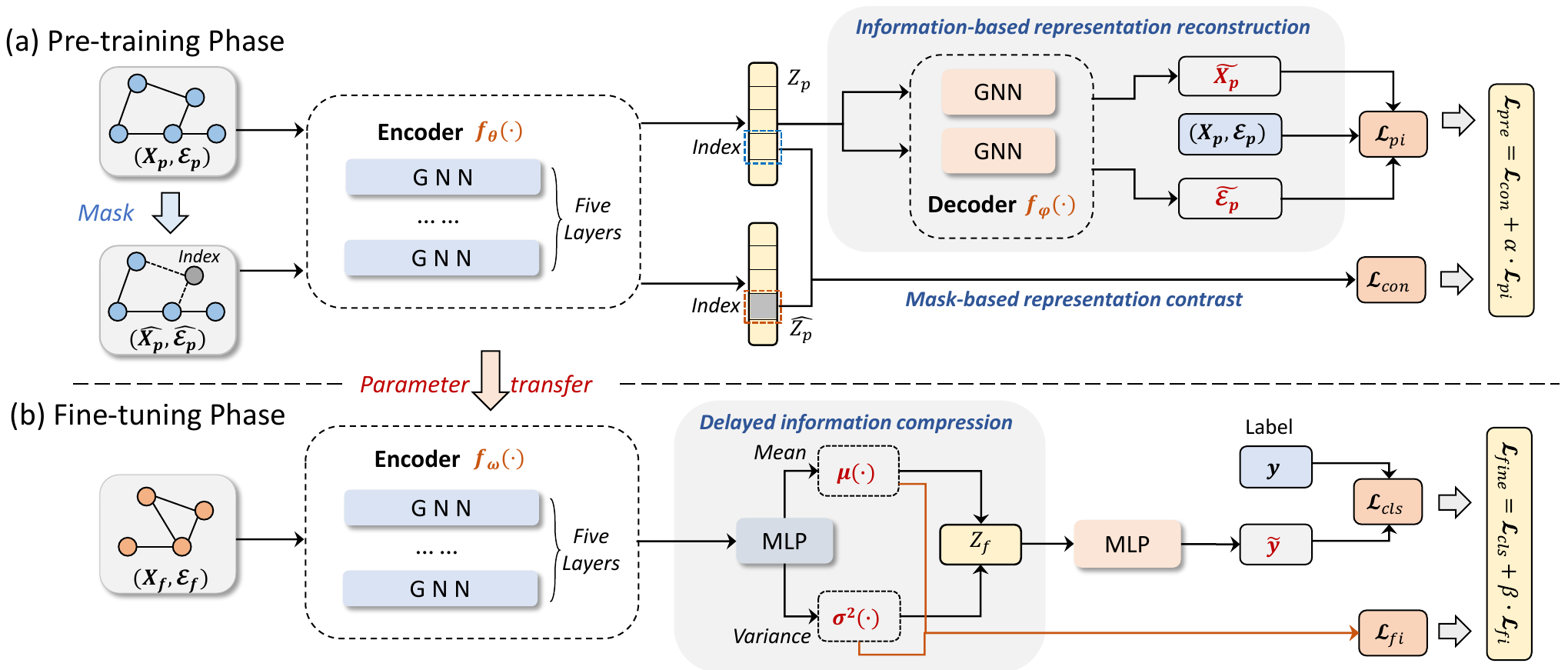}
    \caption{Architecture of DBP framework. Subfigure (a) corresponds to the generative and contrastive learning based self-supervised pre-training model. The optimization objective of pre-training consists of $L_{con}$ and $ L_{pi}$ which are respectively used to extract general knowledge and avoid excessive information compression. Subfigure (b) indicates the information control based fine-tuning model. The optimization objective of fine-tuning, which is composed of $L_{cls}$ and $L_{fi}$, encourages enhanced information compression to improve classification performance. The two-phase transition is implemented by means of parameter transfer.}
    \label{fig:framework}
\end{figure*}

\subsection{Delayed Bottlenecking Strategy}
In this subsection, we propose a novel strategy, Delayed Bottlenecking (DBP), to alleviate the above problem. The basic idea of such a strategy is to suppress the information compression imposed for the pre-training task in pre-training and enhance the compression based on the downstream task in fine-tuning, i.e., make sure $I_{\theta_{0}}(f_{\theta_{0}};\mathcal{D}^{pre})$ be closer to $I_{\theta_{max}}(f_{\theta_{max}};\mathcal{D}^{pre})$ while obtaining the optimal pre-training parameter set $\theta_0$. 

Such operation of maintaining as much as possible information to fine-tuning phase can also be viewed as that the pre-trained parameters and representation are skewed to the downstream task, and to achieve this target, we can formulate two information control objectives respectively for pre-training and fine-tuning phases, i.e.,

\textit{i) Pre-training phase:}
\begin{equation}\label{eq:1}
    \mathcal{L}_{pi} = - I(\mathcal{D}^{pre}; Z)
\end{equation}

\textit{ii) Fine-tuning phase:}
\begin{equation}\label{eq:2}
\begin{small}
\begin{matrix}
 \mathcal{L}_{fine}   = \mathcal{L}_{cls} + \beta \cdot \mathcal{L}_{fi}\\
\\
\;\;\;\;\;\;\;\;\;\;\;\; where\;\;\;\;
\left\{\begin{matrix}
 \mathcal{L}_{cls}=-I(Y; Z)\\
 \\
\mathcal{L}_{fi}=I(\mathcal{D}^{fine}; Z)
\end{matrix}\right.
\end{matrix}
\end{small}
\end{equation}
where $I(\cdot; \cdot)$ represents the mutual information, $Z$ is the latent representation, $Y$ is the downstream target, $\beta$ is employed to control the degree of enhancing compression in fine-tuning and can be tuned based on task and dataset. $\mathcal{L}_{fine}$ is used to enhance compression with the guidance of the downstream task and labeled fine-tuning data. Therefore, it contains two components, $\mathcal{L}_{fi}$ is to enhance the compression based on labeled fine-tuning data, and $\mathcal{L}_{cls}$ is to enhance the mutual information between latent representation and downstream task, hence improving the final performance of our model in the downstream task.  Different from traditional optimization problems, in the field of deep learning, the issue of optimizing such objectives can also be converted into seeking the variational upper bounds respectively for $\mathcal{L}_{pi}$ and $\mathcal{L}_{fine}$ to achieve the minimization constraint of mutual information in both pre-training and fine-tuning periods. We then detailedly discuss these in the next subsection.

\subsection{Information Control Objectives for Optimization}\label{sec:3.3}
Due to the intractability of mutual information, the IB objectives in Equations~\ref{eq:1} and~\ref{eq:2} are hard to be directly used in optimization. Therefore, in this subsection, we derive the tractable upper bounds of $\mathcal{L}_{pi}$ and $\mathcal{L}_{fi}$. The variational upper bounds ensure that the original mutual information objective can be reduced in case the empirical risks of $\mathcal{L}_{fine}$ and $\mathcal{L}_{pi}$ are reduced.
\begin{proposition}[\textbf{Upper bound of $\mathcal{L}_{pi}$}]\label{pro1}
Given pre-training dataset $\mathcal{D}^{pre}$, latent representation $Z_p$ learned from $\mathcal{D}^{pre}$, and graph $G_p=(X_p,\mathcal{E}_p) \in \mathcal{D}^{pre}$, we have
\begin{equation}
    \begin{aligned}
       \mathcal{L}_{pi} &= -I(\mathcal{D}^{pre}; Z) 
       \\
       &\leq -\mathbb{E}_{Z_p \sim p_\theta {(Z_p|X_p,\mathcal{E}_p)} } [\mathrm{log}\;q_\varphi (X_p,\mathcal{E}_p|Z_p)]
    \end{aligned}\label{eq:5}
\end{equation}
where $p_\theta {(Z_p|X_p,\mathcal{E}_p)}$ is the variational approximation of true conditional probability $p(Z_p|X_p,\mathcal{E}_p)$ in the encoder during pre-training, and $q_\varphi(X_p,\mathcal{E}_p|Z_p)$ is the variational approximation of the true conditional probability $q(X_p,\mathcal{E}_p|Z_p)$ in the decoder during pre-training.
\end{proposition}
\noindent\textit{\textbf{Proof.}} For pre-training dataset $\mathcal{D}^{pre}$, latent representation $Z_p$ learned from $\mathcal{D}^{pre}$, and graph $G_p=(X_p,\mathcal{E}_p) \in \mathcal{D}^{pre}$, we have:
\begin{small}
    \begin{equation}\label{app:eq2}
    \begin{aligned}
       I(\mathcal{D}^{pre}; Z_p) & = H(\mathcal{D}^{pre}) - H(\mathcal{D}^{pre}|Z_p) \\
       & \ge - H(\mathcal{D}^{pre}|Z_p) \\
       & \overset{(1)}{=}  \int d Z_p d G_p \, p(Z_p, G_p) \, \mathrm{log} \, p(G_p|Z_p) \\
       & \overset{(2)}{=}  \int d Z_p d G_p \, p(Z_p, G_p) \, \mathrm{log} \, q_\varphi(G_p|Z_p) \\
       & \quad \; + \int d Z_p d G_p \, p(Z_p, G_p) \, \mathrm{log} \, \frac {p(G_p|Z_p)} {q_\varphi(G_p|Z_p)} \\
       & \overset{(3)}{=}  \int d Z_p d G_p \, p(Z_p, G_p) \, \mathrm{log} \, q_\varphi(G_p|Z_p) \\
       & \quad \; + \int d G_p \, p(G_p|Z_p) \, \mathrm{log} \, \frac {p(G_p|Z_p)} {q_\varphi(G_p|Z_p)} \\
       & \overset{(4)}{=}  \int d Z_p \, p_\theta(Z_p|G_p) \, \mathrm{log} \, q_\varphi(G_p|Z_p) \\
       & \quad \; + \mathrm{KL} \, [p(G_p|Z_p) || q(G_p|Z_p)] \\
       & \overset{(5)}{\ge} \int d Z_p \, p_\theta(Z_p|G_p) \, \mathrm{log} \, q_(G_p|Z_p) \\
       & \overset{(6)}{=}  \mathbb{E}_{Z_p \sim p_\theta {(Z_p|G_p)} } [\mathrm{log}\,q_\varphi (G_p|Z_p)] \\
       & \overset{(7)}{=}  \mathbb{E}_{Z_p \sim p_\theta {(Z_p|X_p,\mathcal{E}_p)} } [\mathrm{log}\,q_\varphi (X_p,\mathcal{E}_p|Z_p)].
    \end{aligned}
\end{equation}
\end{small}

\noindent Among Equation \ref{app:eq2}, step (1) is the definition of mutual information. Steps (2) and (3) are based on the integral property. Steps (4) and (5) are defined according to the KL divergence. Steps (6) and (7) are based on the properties of integrals and expectations. 
\begin{equation}\label{app:eq3}
    \begin{aligned}
       \mathcal{L}_{pi} &= -\alpha\cdot I(\mathcal{D}^{pre}; Z_p) 
       \\
       & \leq -\alpha\cdot \mathbb{E}_{Z_p \sim p_\theta {(Z_p|X_p,\mathcal{E}_p)} } [\mathrm{log}\,q_\varphi (X_p,\mathcal{E}_p|Z_p)]
    \end{aligned}
\end{equation}

\noindent In the fine-tuning phase, we encourage the downstream task can compress information with ground-truth labels, so that the learned knowledge during pre-training can be transferred and generalized on the downstream task more quickly.
\begin{proposition}[\textbf{Upper bound of $\mathcal{L}_{fine}$}]\label{pro2}
Given fine-tuning dataset $\mathcal{D}^{fine}$, latent representation $Z_f$ learned from $\mathcal{D}^{fine}$, the label $y$ of downstream task $Y$, and graph $G_f=(X_f,\mathcal{E}_f) \in \mathcal{D}^{fine}$, we have 
\begin{equation}
    \begin{aligned}
        \mathcal{L}_{fine} = \, & \beta \cdot I(\mathcal{D}^{fine}; Z) - I(Y; Z)
        \\
        \leq \, &  \beta \cdot \mathbb{E}_{Z_f \sim p_\omega {(Z_f|X_f,\mathcal{E}_f)}} \, \mathrm{KL}[p_\omega(Z_f|X_f,\mathcal{E}_f), r(Z_f)]
        \\
        & - \mathbb{E}_{Z_f \sim p_\omega {(Z_f|X_f,\mathcal{E}_f)}}[log\;q_\gamma(y|Z_f)]
    \end{aligned}
\end{equation}
\noindent where $p_\omega {(Z_f|X_f,\mathcal{E}_f)}$ is the variational approximation of true conditional probability $p {(Z_f|X_f,\mathcal{E})_f}$ in the encoder during fine-tuning, $r(Z_f)$ is an estimation of prior probability $p(Z_f)$ of $Z_f$, and $q_\gamma(y|Z_f)$ is the variational approximation of the true conditional probability $q(y|Z_f)$ in the classifier. 
\end{proposition}
\noindent\textit{\textbf{Proof.}} For fine-tuning dataset $\mathcal{D}^{fine}$, latent representation $Z_f$ learned from $\mathcal{D}^{fine}$, the label of downstream task $y$, and graph $G_f=(X_f,\mathcal{E}_f) \in \mathcal{D}^{fine}$, we have:
\begin{small}
    \begin{equation}\label{app:eq5}
    \begin{aligned}
        & I(y; Z_f) - \beta \cdot I(\mathcal{D}^{fine}; Z_f) \\
        & \ge \int d X_f d y d Z_f \, p(X_f) \, p(y|X_f) \, p(Z_f|X_f) \, \mathrm{log} \, q(y|Z_f) \\
        & \quad \; - \beta \cdot \int d X_f d Z_f \, p(X_f) \, p(Z_f|X_f) \, \mathrm{log} \, \frac{p(Z_f|X_f)}{r(Z_f)} \\
        & \overset{(1)}{=} \int d Z_f \, p_\omega(Z_f|X_f) \, \mathrm{log} \, q_\gamma(y|Z_f) \\
        & \quad \; - \beta \cdot \int d Z \, p_\omega(Z_f|X_f) \, \mathrm{log} \, \frac{p_\omega(Z_f|X_f)}{r(Z_f)} \\
        & \overset{(2)}{=} \mathbb{E}_{Z_f \sim p_\omega {(Z_f|X_f,\mathcal{E}_f)}}[log\;q_\gamma(y|Z_f)] \\
        & \quad \; - \beta \cdot \mathbb{E}_{Z_f \sim p_\omega {(Z_f|X_f,\mathcal{E}_f)}} \, \mathrm{KL}[p_\omega(Z_f|X_f,\mathcal{E}_f), r(Z_f)].
    \end{aligned}
\end{equation}
\end{small}

\vspace{0.1in}

\noindent Among Equation \ref{app:eq5}, step (1) is based on the properties of conditional probability and marginal probability. Step (2) follows from the definition of expectation. Thus, we can obtain an upper bound on the information control objective of the fine-tuning stage:
   \begin{equation}
    \begin{aligned}
        \mathcal{L}_{fine} = \, & \beta \cdot I(\mathcal{D}^{fine}; Z_f) - I(y; Z_f)
        \\
        \leq \, &  \beta \cdot \mathbb{E}_{Z_f \sim p_\omega {(Z_f|X_f,\mathcal{E}_f)}} \, \mathrm{KL}[p_\omega(Z_f|X_f,\mathcal{E}_f), r(Z_f)]
        \\
        & - \mathbb{E}_{Z_f \sim p_\omega {(Z_f|X_f,\mathcal{E}_f)}}[log\;q_\gamma(y|Z_f)].
    \end{aligned}
\end{equation} 

\subsection{Proof for Parameters Transfer}
In this section, we will prove how the proposed two-stage loss function improves the transfer of pre-trained parameters. First, we introduce some additional definitions and lemmas:

\begin{definition}[KL Divergence]
For two probability distributions $P$ and $Q$, the KL divergence between them is defined as:

\begin{equation}
    D_{\text{KL}}(P \| Q) = \mathbb{E}_{x \sim P}\left[ \log \frac{P(x)}{Q(x)}\right]
\end{equation}
\end{definition}

\begin{lemma}[Chain Rule of KL Divergence]\label{lemma:chain_rule_kl}
For three probability distributions \( P(X,Y) \), \( Q(X,Y) \), and \( R(X) \), we have:
\begin{equation}
\begin{split}
    D_{\text{KL}}(P(X,Y) \| & Q(X,Y)) = \,  D_{\text{KL}}(P(X) \| R(X)) \\
    & + \mathbb{E}_{x \sim P(X)}[D_{\text{KL}}(P(Y|X) \| Q(Y|X))]
\end{split}
\end{equation}
\end{lemma}


\begin{lemma}[Non-Negativity of KL Divergence]\label{lemma:kl_nonnegative}
For any two probability distributions $P$ and $Q$, we have $D_{\text{KL}}(P \| Q) \geq 0$, with equality holding if and only if $P=Q$ almost everywhere.
\end{lemma}



The equality holds if and only if $\frac{Q(x)}{P(x)}$ is a constant almost everywhere, i.e., $P=Q$ almost everywhere. The proof of this lemma can be demonstrated using Jensen's inequality, but is omitted here due to space constraints.

Now, we restate and prove the main theorem:

\begin{theorem}[Bounding Posterior Distributions via DBP]\label{thm:bayes_refined}
Let $\mathcal{D}^{pre}, \mathcal{D}^{fine}$ denote the pre-training data and fine-tuning data, respectively, $Z_p, Z_f$ denote the corresponding latent representations, and $Y$ denote the labels for the downstream task. Define:

\begin{equation}
\left\{\begin{matrix}
\mathcal{L}_{pi} = -I(\mathcal{D}^{pre}; Z_p) \;\;\;\;\;\;\;\;\;\;\;\;\;\;\;\;\;\;\;\;\\
 \\
\mathcal{L}_{fi} = \beta I(\mathcal{D}^{fine}; Z_f) - I(Y; Z_f)
\end{matrix}\right.
\end{equation}
We make the following assumptions:
\begin{assumption}\label{ass1}
$H$ denotes the model hypothesis space, $P(H), Q(H)$ denote two prior distributions;
\end{assumption}
\begin{assumption}\label{ass2}
$P(Z_p|\mathcal{D}^{pre}, H) = Q(Z_p|\mathcal{D}^{pre}, H), \forall H \in \mathcal{H}$;
\end{assumption}
\begin{assumption}\label{ass3}
$P(Z_f|\mathcal{D}^{fine}, H) = Q(Z_f|\mathcal{D}^{fine}, H), \forall H \in \mathcal{H}$;
\end{assumption}
\begin{assumption}\label{ass4}
$P(Y|Z_f, H) = Q(Y|Z_f, H), \forall H \in \mathcal{H}$.
\end{assumption}
Then the optimization objectives $\mathcal{L}_{pi}$ and $\mathcal{L}_{fi}$ satisfy:
\begin{equation}
    \begin{aligned}
    &D_{\text{KL}}(P(H|\mathcal{D}^{pre}) \| Q(H|\mathcal{D}^{pre})) \\
    &\quad\leq D_{\text{KL}}(P(H) \| Q(H)) - \mathcal{L}_{pi}(P) + \mathcal{L}_{pi}(Q) \\
    &D_{\text{KL}}(P(H|\mathcal{D}^{fine}) \| Q(H|\mathcal{D}^{fine})) \\
    &\quad\leq D_{\text{KL}}(P(H) \| Q(H)) + \frac{1}{\beta}(\mathcal{L}_{fi}(P) - \mathcal{L}_{fi}(Q))
\end{aligned}
\end{equation}
\end{theorem}
Theorem \ref{thm:bayes_refined} provides a theoretical basis for improving the parameter transfer process by indicating that minimizing $\mathcal{L}_{pi}$ and $\mathcal{L}_{fi}$ can control the difference between the posterior distributions of the prior distribution under the pre-training and fine-tuning data, respectively. Specifically:
\begin{itemize}
    \item Minimizing $\mathcal{L}_{pi}(P)$ aligns the posterior $P(H|\mathcal{D}^{pre})$ after pre-training with the reference posterior $Q(H|\mathcal{D}^{pre})$, preventing overfitting to the pre-training data when $Q(H)$ is the prior. This suppresses excessive information compression during pre-training, allowing the learned parameters and representations to adequately retain information from the pre-training data for subsequent fine-tuning.
    \item Minimizing $\mathcal{L}_{fi}(P)$ aligns the posterior $P(H|\mathcal{D}^{fine})$ after fine-tuning with $Q(H|\mathcal{D}^{fine})$, preventing overfitting to the fine-tuning data. Simultaneously, it maximizes $I_{P}(Y; Z_f) - \beta I_{P}(\mathcal{D}^{fine}; Z_f)$, consistent with maximizing the mutual information between the latent representation and downstream labels via $\mathcal{L}_{cls}$ while enhancing information compression via $\mathcal{L}_{fi}$ to adapt the pre-trained parameters to the downstream task during fine-tuning. This reflects the core delayed bottlenecking strategy of delaying the focus of information compression to the fine-tuning stage, guided by downstream supervision, which can better control parameter learning in both phases compared to simply maximizing pre-training mutual information.
\end{itemize} 
\noindent\textit{\textbf{Proof.}} First, we prove the first inequality. By Lemma \ref{lemma:chain_rule_kl}, we have:
    \begin{equation}
        \begin{aligned}
            &D_{\text{KL}}(P(H,Z_p,\mathcal{D}^{pre}) \| Q(H,Z_p,\mathcal{D}^{pre})) \\
            &= D_{\text{KL}}(P(H) \| Q(H))  \\
            &+ \mathbb{E}_{ P(H)}[D_{\text{KL}}(P(Z_p,\mathcal{D}^{pre}|H) \| Q(Z_p,\mathcal{D}^{pre}|H))] \\
            &= D_{\text{KL}}(P(H) \| Q(H))  \\
            &+ \mathbb{E}_{P(H)}[D_{\text{KL}}(P(\mathcal{D}^{pre}|H) \| Q(\mathcal{D}^{pre}|H))] \\
            &\quad + \mathbb{E}_{ P(H),  P(\mathcal{D}^{pre}|H)}[D_{\text{KL}}(P(Z_p|\mathcal{D}^{pre},H) \| Q(Z_p|\mathcal{D}^{pre},H))] \\
            &= D_{\text{KL}}(P(H) \| Q(H)) \\
            & + \mathbb{E}_{P(H)}[D_{\text{KL}}(P(\mathcal{D}^{pre}|H) \| Q(\mathcal{D}^{pre}|H))] \\
            &\geq D_{\text{KL}}(P(H) \| Q(H))  \\
            &+ \mathbb{E}_{  P(H)}[D_{\text{KL}}(P(\mathcal{D}^{pre}|H) \| Q(\mathcal{D}^{pre}|H))]
        \end{aligned}
    \end{equation}

\noindent where the second equality uses assumption (2), and the last inequality uses Lemma \ref{lemma:kl_nonnegative}.
On the other hand, we have:
    \begin{equation}
    \begin{aligned}
    &D_{\text{KL}}(P(H,Z_p,\mathcal{D}^{pre}) \| Q(H,Z_p,\mathcal{D}^{pre})) \\
    &\quad = \mathbb{E}_{(  P(H,Z_p,\mathcal{D}^{pre})}\left[ \log \frac{P(h,z_p,d^{pre})}{Q(h,z_p,d^{pre})}\right] \\
    &\quad = \mathbb{E}_{(  P(H,Z_p,\mathcal{D}^{pre})}\left[ \log \frac{P(h)P(d^{pre}|h)P(z_p|d^{pre},h)}{Q(h)Q(d^{pre}|h)Q(z_p|d^{pre},h)}\right] \\
    &\quad = \mathbb{E}_{(  P(H,Z_p,\mathcal{D}^{pre})}\left[ \log \frac{P(h)}{Q(h)} \right. \\
    &\quad\quad \left. + \log \frac{P(d^{pre}|h)}{Q(d^{pre}|h)} + \log \frac{P(z_p|d^{pre},h)}{Q(z_p|d^{pre},h)}\right] \\
    &\quad = D_{\text{KL}}(P(H) \| Q(H)) \\
    &\quad\quad + \mathbb{E}_{    P(H,\mathcal{D}^{pre})}\left[ \log \frac{P(d^{pre}|h)}{Q(d^{pre}|h)}\right] \\
    &\quad\quad + \mathbb{E}_{(  P(H,Z_p,\mathcal{D}^{pre})}\left[ \log \frac{P(z_p|d^{pre},h)}{Q(z_p|d^{pre},h)}\right] \\
    &\quad = D_{\text{KL}}(P(H) \| Q(H)) + I_{P}(\mathcal{D}^{pre}; H) - I_{Q}(\mathcal{D}^{pre}; H) \\
    &\quad\quad + \mathbb{E}_{    P(H,\mathcal{D}^{pre})}[D_{\text{KL}}(P(Z_p|\mathcal{D}^{pre},H) \| Q(Z_p|\mathcal{D}^{pre},H))] \\
    &\quad = D_{\text{KL}}(P(H) \| Q(H)) - \mathcal{L}_{pi}(P) + \mathcal{L}_{pi}(Q)\label{eq:kl_inequality0}
    \end{aligned}
\end{equation}

\noindent Combining the above two inequalities, we obtain:  
\begin{align}
&D_{\text{KL}}(P(H|\mathcal{D}^{pre}) \| Q(H|\mathcal{D}^{pre})) \nonumber \\
&\leq D_{\text{KL}}(P(H,Z_p,\mathcal{D}^{pre}) \| Q(H,Z_p,\mathcal{D}^{pre})) \nonumber \\
&\leq D_{\text{KL}}(P(H) \| Q(H)) - \mathcal{L}_{pi}(P) + \mathcal{L}_{pi}(Q)\label{eq:kl_inequality1}
\end{align}
Next, we prove the second inequality. Similarly, we have:
    \begin{equation}
    \begin{aligned}
    D_{\text{KL}}&(P(H,Z_f,\mathcal{D}^{fine},Y) \| Q(H,Z_f,\mathcal{D}^{fine},Y)) \\
    &= D_{\text{KL}}(P(H) \| Q(H))\\
    &\quad + \mathbb{E}_{P(H)}\Big[D_{\text{KL}}(P(Z_f,\mathcal{D}^{fine},Y|H) \| \\
    &Q(Z_f,\mathcal{D}^{fine},Y|H))\Big] \\
    &= D_{\text{KL}}(P(H) \| Q(H))  \\
    &\quad + \mathbb{E}_{P(H)}\Big[D_{\text{KL}}(P(\mathcal{D}^{fine}|H)\| Q(\mathcal{D}^{fine}|H))\Big]\\
    &\quad + \mathbb{E}_{P(H)}\mathbb{E}_{P(\mathcal{D}^{fine}|H)}\Big[D_{\text{KL}}(P(Z_f|\mathcal{D}^{fine},H)  \\
    &\qquad \| Q(Z_f|\mathcal{D}^{fine},H))\Big]  \\
    &\quad + \mathbb{E}_{P(H)}\mathbb{E}_{P(\mathcal{D}^{fine}|H)}\mathbb{E}_{P(Z_f|\mathcal{D}^{fine},H)}\Big[D_{\text{KL}}(P(Y|Z_f,H)  \\
    &\qquad \| Q(Y|Z_f,H))\Big]  \\
    &= D_{\text{KL}}(P(H) \| Q(H))  + \mathbb{E}_{P(H)}\Big[D_{\text{KL}}(P(\mathcal{D}^{fine}|H)  \\
    &\qquad \| Q(\mathcal{D}^{fine}|H))\Big]  \\
    &\geq D_{\text{KL}}(P(H) \| Q(H))  \\
    &\quad + \mathbb{E}_{P(H)}\Big[D_{\text{KL}}(P(\mathcal{D}^{fine}|H)\| Q(\mathcal{D}^{fine}|H))\Big] \label{eq:kl_inequality2}
    \end{aligned}
\end{equation}
\noindent where the third equality uses Assumptions \ref{ass3} and \ref{ass4}, and the last inequality uses Lemma \ref{lemma:kl_nonnegative}.
On the other hand, we have:
\begin{equation}
    \begin{aligned}
&D_{\text{KL}}(P(H,Z_f,\mathcal{D}^{fine},Y) \| Q(H,Z_f,\mathcal{D}^{fine},Y)) \\
&=\mathbb{E}_{(  P(H,Z_f,\mathcal{D}^{fine},Y)}\bigg[ \log \frac{P(h,z_f,d^{fine},y)}{Q(h,z_f,d^{fine},y)}\bigg] \\
&= \mathbb{E}_{(     P(H,Z_f,\mathcal{D}^{fine},Y)}\bigg[ \log \frac{P(h)}{Q(h)} + \log \frac{P(d^{fine}|h)}{Q(d^{fine}|h)} \\
&\quad + \log \frac{P(z_f|d^{fine},h)}{Q(z_f|d^{fine},h)} + \log \frac{P(y|z_f,h)}{Q(y|z_f,h)}\bigg] \\
&= D_{\text{KL}}(P(H) \| Q(H)) + \mathbb{E}_{  P(H,\mathcal{D}^{fine})}\bigg[ \log \frac{P(d^{fine}|h)}{Q(d^{fine}|h)}\bigg] \\
&\quad + \mathbb{E}_{   P(H,Z_f,\mathcal{D}^{fine})}\bigg[ \log \frac{P(z_f|d^{fine},h)}{Q(z_f|d^{fine},h)}\bigg] \\
&\quad + \mathbb{E}_{ P(H,Z_f,Y)}\bigg[ \log \frac{P(y|z_f,h)}{Q(y|z_f,h)}\bigg] \\
&= D_{\text{KL}}(P(H) \| Q(H)) + I_{P}(\mathcal{D}^{fine}; H) - I_{Q}(\mathcal{D}^{fine}; H) \\
&\quad + I_{P}(Z_f; \mathcal{D}^{fine}|H) - I_{Q}(Z_f; \mathcal{D}^{fine}|H) \\
&\quad + I_{P}(Y; Z_f|H) - I_{Q}(Y; Z_f|H) \\
&= D_{\text{KL}}(P(H) \| Q(H)) + \beta(I_{P}(\mathcal{D}^{fine}; Z_f) - I_{Q}(\mathcal{D}^{fine}; Z_f)) \\
&\quad - (I_{P}(Y; Z_f) - I_{Q}(Y; Z_f)) \\
&= D_{\text{KL}}(P(H) \| Q(H)) + \frac{1}{\beta}(\mathcal{L}_{fi}(P) - \mathcal{L}_{fi}(Q))
\end{aligned}
\end{equation}
Combining the above two inequalities, we obtain:
\begin{align}
&D_{\text{KL}}(P(H|\mathcal{D}^{fine}) \| Q(H|\mathcal{D}^{fine}))\nonumber \\
&\leq D_{\text{KL}}(P(H,Z_f,\mathcal{D}^{fine},Y) \| Q(H,Z_f,\mathcal{D}^{fine},Y))\nonumber \\
&\leq D_{\text{KL}}(P(H) \| Q(H)) + \frac{1}{\beta}(\mathcal{L}_{fi}(P) - \mathcal{L}_{fi}(Q))
\end{align}
This completes the proof of the theorem.

\section{DBP Framework}

Given the theoretical analysis in the previous section, we still have to consider how to integrate the proposed DBP strategy into the actual model design, so that the information control objectives can be applied to graph structure data. We will discuss these issues in this section.
\subsection{Solution Overview}

The solution overview is illustrated in Figure \ref{fig:framework} which also contains the implementations of the above two optimization information control objectives in the pre-training and fine-tuning phases. The DBP framework contains two parts: i) a self-supervised pre-training model for controlling information and extracting knowledge via generative learning and contrastive learning, and ii) a fine-tuning model with an information control module.

\subsection{Generative and Contrastive Learning Based Self-supervised Pre-training}
As shown in Figure \ref{fig:framework}(a), We built a self-supervised pre-training model that consists of two components, mask-based representation contrast and information-based representation reconstruction. The first one is to extract general knowledge from pre-training data, and the second component is used to maintain the mutual information between latent representation and training data. We will describe these two modules in detail in the following subsections.

\noindent \textbf{Mask-based Representation Contrast.}
Before pre-training, we first use a random masking scheme to mask some nodes and their connected edges in the original graph $G_p=(X_p,\mathcal{E}_p)$ and record the indices of the masked nodes to obtain the noisy graph $\widehat{G_p} = (\widehat{X_p},\widehat{\mathcal{E}_p})$. Then, the original graph $G_p$ and the noisy graph $\widehat{G_p}$ are used as the input of encoder $f_\theta$ to generate the original graph node representation $Z_p$ and the noisy graph node representation $\widehat{Z_p}$, i.e.,
\begin{equation}
    Z_p = f_\theta(X_p,\mathcal{E}_p), \quad \widehat{Z_p} = f_\theta(\widehat{X_p},\widehat{\mathcal{E}_p})
\end{equation}
\noindent To realize self-supervised learning, we define the product between the representations of masked nodes and their connected nodes in the noisy graph as the negative sample, while the product of representations in the original graph is correspondingly defined as the positive sample. The contrastive self-supervised objective can be written as,
\begin{small}
    \begin{equation}
    \mathcal{L}_{con} = \sum_{u,u' \in mask} -\mathrm{ln}(\sigma({Z_p}_u^T \cdot {Z_p}_v)) - \mathrm{ln}(\sigma(-\widehat{{Z_p}_{u'}}^T \cdot \widehat{{Z_p}_{v'}}))
\end{equation}
\end{small}

\noindent where $u$ and $v$ are correspondingly the masked nodes and their connected nodes in the original graph, while $u'$ and $v'$ are the masked nodes and their connected nodes in the noisy graph. Note here $\sigma$ indicates the $\mathrm{Sigmoid}$ function. 

\noindent \textbf{Information-Maintain Representation Reconstruction.}
We further introduce the information control theory mentioned in Proposition \ref{pro1} into the node representation learning of the original graph. Here, to obtain the representation $Z_p$ of the original graph by encoder $f_\theta$, we let the conditional probability $p_\theta (Z_p|X_p,\mathcal{E}_p)$ be the variational approximation of the true conditional probability $p(Z_p|X_p,\mathcal{E}_p)$. To calculate $q_\varphi(X_p,\mathcal{E}_p|Z_p)$, we employ a decoder $f_\varphi$ consisting of two single-layer GNNs to reconstruct node features $\widetilde{X_p}$ and edge features $\widetilde{\mathcal{E}_p}$ of the original graph from $Z_p$, i.e., 
\begin{equation}
    \widetilde{X_p}, \widetilde{\mathcal{E}_p} = f_\varphi (Z_p)
\end{equation}
The optimization goal of our information reconstruction task can be calculated as the cross-entropy (CE) loss between reconstructed features and original features, and it can be simplified as the upper bound derived in Equation \ref{eq:5}.
\begin{equation}
    \mathcal{L}_{pi} = - \mathbb{E}_{Z_p\sim p_\theta {(Z_p|X_p,\mathcal{E}_p)} } [log\;q_\varphi (X_p,\mathcal{E}_p|Z_p)]
\end{equation}
\noindent Here, $\mathcal{L}_{pi}$ allows the representation to retain more information from the pre-trained dataset by encouraging the reconstructed nodes and features to have more similarity to the original graph. It is not only an approximation of our proposed information control objective, but also indicates that the latent representation encoded by $f_\theta$ has the potential to be restored to the original representation. 

\noindent \textbf{Pre-training Objective.} The above representation contrast and reconstruction components work jointly to extract general knowledge from pre-training data and simultaneously suppress information compression in latent representation learning. This jointly working mechanism determines that the pre-trained model can effectively avoid information forgetting in extracting general knowledge, and information compression is delayed to the fine-tuning phase. Therefore, the overall loss of pre-training should be the sum of the objectives of these two components, i.e.,
    \begin{equation}
\begin{aligned}
    \mathcal{L}_{pre} & = \mathcal{L}_{con} + \alpha \cdot \mathcal{L}_{pi} \\
    & = \sum_{u,u' \in mask} -\mathrm{ln}(\sigma({Z_p}_u^T \cdot {Z_p}_v)) - \mathrm{ln}(\sigma(-\widehat{{Z_p}_{u'}}^T \cdot \widehat{{Z_p}_{v'}})) \\ 
    & \quad \; - \alpha \cdot \mathbb{E}_{Z_p\sim p_\theta {(Z_p|X_p,\mathcal{E}_p)} } [log\;q_\varphi (X_p,\mathcal{E}_p|Z_p)]
\end{aligned}
\end{equation}
\noindent where $\alpha$ is a hyper-parameter for adjusting the weight of information control.

\subsection{Information Controlled Fine-tuning}
Similar to recent studies, during fine-tuning, we also employ the same encoder $f_\omega$, whose parameters $\omega$ are initialized with the parameter $\theta$ of $f_\theta$, as in the pre-training phase. Notice that the labeled map $G_f = (X_f,\mathcal{E}_f)$ in downstream task is encoded to obtain the latent representation $Z_f$ by $f_\omega$ which calculates the modeling probability $p_\omega (Z_f|X_f,\mathcal{E}_f)$, i.e.,
\begin{equation}
    Z_f = f_\omega(X_f, \mathcal{E}_f).
\end{equation}
As demonstrated in Figure \ref{fig:framework}(b), different from other pre-training and fine-tuning works, to avoid information forgetting, the information compression in pre-training is suppressed, therefore, we should enhance this operation in the fine-tuning phase. 

\noindent \textbf{Delayed Information Compression.} To achieve delayed information compression in fine-tuning, we add a compression module consisting of two MLPs between $f_\omega$ and the classifier to learn the mean $\mu$ and variance $\sigma^2$ respectively. 
\begin{equation}
   \left\{ \mu, \sigma^2\right \} = \mathrm{MLPs}(Z_f)
\end{equation}
Then we sample the graph representation $Z_f$ from the multivariate normal distribution $\mathcal{N}(Z_f|\mu,\sigma^2)$ with a reparameterization trick to realize back-propagation, i.e., 
\begin{equation}
    Z_t = \mu+\sigma^2\odot\varepsilon
    \end{equation}
where $\varepsilon \sim \mathcal{N}(0,1)$. We further refer to the information control theory mentioned in Proposition \ref{pro2} to calculate the Kullback-Leibler ($\mathrm{KL}$) divergence of distribution $\mathcal{N}(Z_f|\mu,\sigma^2)$ and Gaussian prior distribution $r(Z_f)$ to realize information control during fine-tuning. The optimization objective is the term in formula (6) to enhance compression,
\begin{equation}
    \mathcal{L}_{fi} = \mathbb{E}_{Z_f \sim p_\omega {(Z_f|X_f,\mathcal{E}_f)}} \, \mathrm{KL}[p_\omega(Z_f|X_f,\mathcal{E}_f), r(Z_f)]
\end{equation}

\noindent \textbf{Fine-tuning Objective.} Furthermore, we first take $Z_f$ as the input of classifier $f_\gamma$ to compute the variational approximation $q_\gamma(y|Z_f)$, and the classification loss is the first term in Equation \ref{eq:2}, 
\begin{equation}
    \mathcal{L}_{cls} = - \mathbb{E}_{Z_f \sim p_\omega {(Z_f|X_f,\mathcal{E}_f)}}[log\;q_\gamma(y|Z_f)].
\end{equation}
And based on this, the overall loss of fine-tuning can be defined as,
\begin{equation}
    \begin{aligned}
        \mathcal{L}_{fine} & = \mathcal{L}_{cls} + \beta \cdot \mathcal{L}_{fi}
        \\
        &  =  - \mathbb{E}_{Z_f \sim p_\omega {(Z_f|X_f,\mathcal{E}_f)}}[log\;q_\gamma(y|Z_f)] +
        \\
        & \quad \; \beta \cdot \mathbb{E}_{Z_f \sim p_\omega {(Z_f|X_f,\mathcal{E}_f)}} \, \mathrm{KL}[p_\omega(Z_f|X_f,\mathcal{E}_f), r(Z_f)],
    \end{aligned}
\end{equation}
where $\beta$ is a hyper-parameter for adjusting the weight of information control.

\begin{table*}[!t]
	\centering
    \caption{ROC-AUC scores (\%) on downstream molecular property prediction task compared with state-of-the-art methods. \textbf{Bold} indicates the best performance while \underline{underline} indicates the second best on each dataset.}
    \label{tbl:MPP}
	\scalebox{1.02}{
		\begin{tabular}{c|  c c c  c  c c c  c | c c}
			\toprule[2pt]
			\textbf{Method}	        & \textbf{BBBP} &\textbf{Tox21} & \textbf{ToxCast} & \textbf{SIDER} & \textbf{ClinTox} & \textbf{MUV} & \textbf{HIV} & \textbf{BACE} & \textbf{AVG.} & \textbf{GAIN}
            \\
			\midrule
            \midrule
			No Pre-training     & 65.8 ± 4.5 & 74.0 ± 0.8 & 63.4 ± 0.6 & 57.3 ± 1.6 & 58.0 ± 4.4 & 71.8 ± 2.5 & 75.3 ± 1.9 & 70.1 ± 5.4 & 67.0 & -  \\
			\midrule
			EdgePred        & 67.3 ± 2.4 & 76.0 ± 0.6 & 64.1 ± 0.6 & 60.4 ± 0.7 & 64.1 ± 3.7 & 74.1 ± 2.1 & 76.3 ± 1.0 & 79.9 ± 0.9 & 70.3 & 3.3\\
			InfoGraph       & 68.2 ± 0.7 & 75.5 ± 0.6 & 63.1 ± 0.3 & 59.4 ± 1.0 & 70.5 ± 1.8 & 75.6 ± 1.2 & \underline{77.6 ± 0.4} & 78.9 ± 1.1 & 71.1 & 4.1 \\
			AttrMasking     & 64.3 ± 2.8 & \underline{76.7 ± 0.4} & \underline{64.2 ± 0.5} & 61.0 ± 0.7 & 71.8 ± 4.1 & 74.7 ± 1.4 & 77.2 ± 1.1 & 79.3 ± 1.6 & 71.1 & 4.1 \\
			ContextPred     & 68.0 ± 2.0 & 75.7 ± 0.7 & 63.9 ± 0.6 & 60.9 ± 0.6 & 65.9 ± 3.8 & 75.8 ± 1.7 & 77.3 ± 1.0 & 79.6 ± 1.2 & 70.9 & 3.9 \\
			GraphPartition  & 70.3 ± 0.7 & 75.2 ± 0.4 & 63.2 ± 0.3 & 61.0 ± 0.8 & 64.2 ± 0.5 & 75.4 ± 1.7 & 77.1 ± 0.7 & 79.6 ± 1.8 & 70.8 & 3.8\\
            GraphCL         & 69.5 ± 0.5 & 75.4 ± 0.9 & 63.8 ± 0.4 & 60.8 ± 0.7 & 70.1 ± 1.9 & 74.5 ± 1.3 & 77.6 ± 0.9 & 78.2 ± 1.2 & 71.3 & 4.3\\
            GraphLoG        & \underline{72.5 ± 0.8} & 75.7 ± 0.5 &  63.5 ± 0.7 & \underline{61.2 ± 1.1 }& 76.7 ± 3.3 & 76.0 ± 1.1 & 77.8 ± 0.8 & \underline{83.5 ± 1.2} & 73.4 & 6.4\\

            GraphMAE        & 72.1 ± 0.5 & 75.2 ± 0.6 &  64.0 ± 0.2 & 60.1 ± 1.1 & 82.1 ± 1.1 &  76.3 ± 2.4 & 76.9 ± 1.0 & 83.1 ± 0.7 & \underline{73.7} & \underline{6.7} \\

            S2GAE           & 71.8 ± 0.5 & 75.5 ± 0.8 & 63.9 ± 0.3 & 60.6 ± 0.8 & \underline{80.7 ± 1.9} & \underline{76.4 ± 1.6} & 76.4 ± 1.5 & 80.5 ± 0.9 & 73.3 & 6.3 \\
            \midrule
             \rowcolor{gray!20}\textbf{DBP}             & \textbf{72.8 ± 0.4} & \textbf{77.8 ± 0.4} & \textbf{65.5 ± 0.3} & \textbf{62.5 ± 0.8} & \textbf{82.8 ± 1.3} & \textbf{77.3 ± 1.1} & \textbf{78.8 ± 1.2} & \textbf{83.7 ± 1.0} & \textbf{74.9} & \textbf{7.9} \\
			\bottomrule[2pt]
		\end{tabular}
	}
\end{table*}

\section{Experiments}
In this section, we compare the performance of our proposed DBP and various state-of-the-art pre-trained baselines on both chemistry and biology domains. Then, we conduct a series of comprehensive model analyses to witness our motivation and the effectiveness of our delayed bottlenecking information control strategy.

\subsection{Experimental Settings}\label{sec:es}
\noindent\textbf{Datasets.} Following the setting of ~\cite{hu2020strategies}, we conduct experiments on data from two domains: molecular property prediction in chemistry and biological function prediction in biology. For chemistry domain, we use Zinc-2M - 2 million unlabeled molecules sampled from the ZINC15 database~\cite{sterling2015zinc} in the pre-training phase and eight binary classification datasets in MoleculeNet~\cite{wu2018moleculenet} in the fine-tuning phase, which are split by the scaffold splitting scheme. For biology domain, we utilize 395K unlabeled protein ego-networks~\cite{hu2020strategies} for self-supervised pre-training and predict 40 fine-grained biological functions of 8 species in the fine-tuning phase. We show the statistics of the two pre-trained datasets and MoleculeNet in Table\ref{tbl:Stam}, respectively.

\begin{table}[!h]
	\centering
    \caption{Statistics for datasets.}
    \label{tbl:Stam}
	\scalebox{0.93}{
		\begin{tabular}{c| c |c c  c c c}
			\toprule[2pt]
			\textbf{Dataset} & \textbf{Type} &\textbf{\#graphs} & \textbf{Avg.\# nodes}& \textbf{\#tasks}\\
			\midrule
            \midrule
            Zinc-2M &Pre-trained & 2,000,000 & 26.6& -\\
	      PPI Networks &Pre-trained & 395,000 & 27.8 & -\\
            \midrule
            BBBP& Chemistry&2,038&24.1&1\\
	      Tox21&Chemistry&7,831&18.6&12\\
            ToxCast&Chemistry&8,598&18.8&617\\
	      SIDER&Chemistry&1,425&33.6&27 \\
            ClinTox&Chemistry&1,478&26.2&2 \\
	      MUV&Chemistry&93,087&14.3&17\\
            HIV&Chemistry&41,127&24.5&1\\
            BACE&Chemistry&1,513&34.1&1\\
            Biological Func. $\times$ 8 & Biology&88,000&-&40\\
			\bottomrule[2pt]
		\end{tabular}
    }
\end{table}

\noindent\textbf{Setups.} Following the setting of~\cite{hu2020strategies}, we employ a five-layer GNN with 300-dimensional hidden units as the encoder and two single-layer GNNs as the decoder in the pre-training phase. We use an Adam optimizer~\cite{kingma2015adam} with a learning rate of $1\times 10^{-3}$ to pre-train the GNN for 100 epochs. In the fine-tuning phase, an information control module and a linear classifier are appended upon the pre-trained GNN and the information control module consists of two MLPs. We also train the model for 100 epochs using the Adam optimizer with learning rate of $1 \times 10^{-3}$ and batch size of 32. We utilize a fixed-step-size learning rate scheduler, which multiplies the learning rate by 0.3 every 30 epochs. All the results are the average of five independent runs with the same configuration and different random seeds.

\noindent\textbf{Attribute Masking Scheme.} On chemistry domain, before pre-training on large-scale molecular graphs, we randomly mask nodes in 25\% of attributes to obtain perturbed graphs. To compare and learn the node representation of the masked position and the node representation of the same position in the original graph, we need to record the index of the mask position for each training data. On biology domain, before pre-training on a large-scale protein self-network graph, we randomly mask the attributes of 25\% of its edges to generate a perturbation graph. We also record the index while controlling the mask by setting additional weights for the variables.

\noindent\textbf{GNN Architectures.} Our experiments are mainly conducted on GIN, but to verify the effectiveness of our method, we conduct experiments with different GNN architectures in Table 3 in the original manuscript. All GNNs in our experiments (e.g., GCN~\cite{DBLP:conf/iclr/KipfW17}, GraphSAGE~\cite{hamilton2017inductive}, GAT~\cite{velivckovic2018graph}, GIN~\cite{DBLP:conf/iclr/XuHLJ19}) are with 5 layers, 300-dimensional hidden units, and a mean
pooling readout function. In addition, two attention heads are employed in each layer of the GAT model.

\noindent\textbf{Baselines.} To demonstrate the effectiveness and robustness of DBP, we compare it with state-of-the-art self-supervised pre-training methods on chemistry and biology domains: EdgePred~\cite{hamilton2017inductive}, InfoGraph~\cite{liu2022unsupervised}, AttrMasking~\cite{hu2020strategies}, ContextPred~\cite{hu2020strategies}, GraphPartition~\cite{you2020does}, GraphCL~\cite{you2020graph}, GraphLoG~\cite{xu2021self}, GraphMAE~\cite{hou2022graphmae}, and S2GAE~\cite{tan2023s2gae}. EdgePred infers link existence between node pairs. ContextPred explores graph structure distribution by sampling and predicting surrounding structures. AttrMasking masks and predicts node or edge attributes based on the neighborhood to learn their distribution. InfoGraph constructs contrastive loss using node representations of the graph instance and other graphs. GraphPartition is a topology-based method that partitions nodes into subsets to minimize cross-subset edges. GraphCL utilizes node dropping, edge perturbation, attribute masking, and subgraph sampling to construct views for contrastive learning. GraphLoG aligns embeddings of related graphs/subgraphs to construct a locally smooth latent space, and models global structure with a hierarchical model. GraphMAE conducts self-supervised pre-training by masking and reconstructing node features, introducing the masked autoencoding idea from computer vision. S2GAE randomly masks edges and learns to reconstruct them with direction-aware masking strategies and a cross-correlation decoder, demonstrating superior performance on link prediction, node classification, and graph classification tasks.
\begin{figure*}[!t]
    \centering  
    \includegraphics[width=1\linewidth]{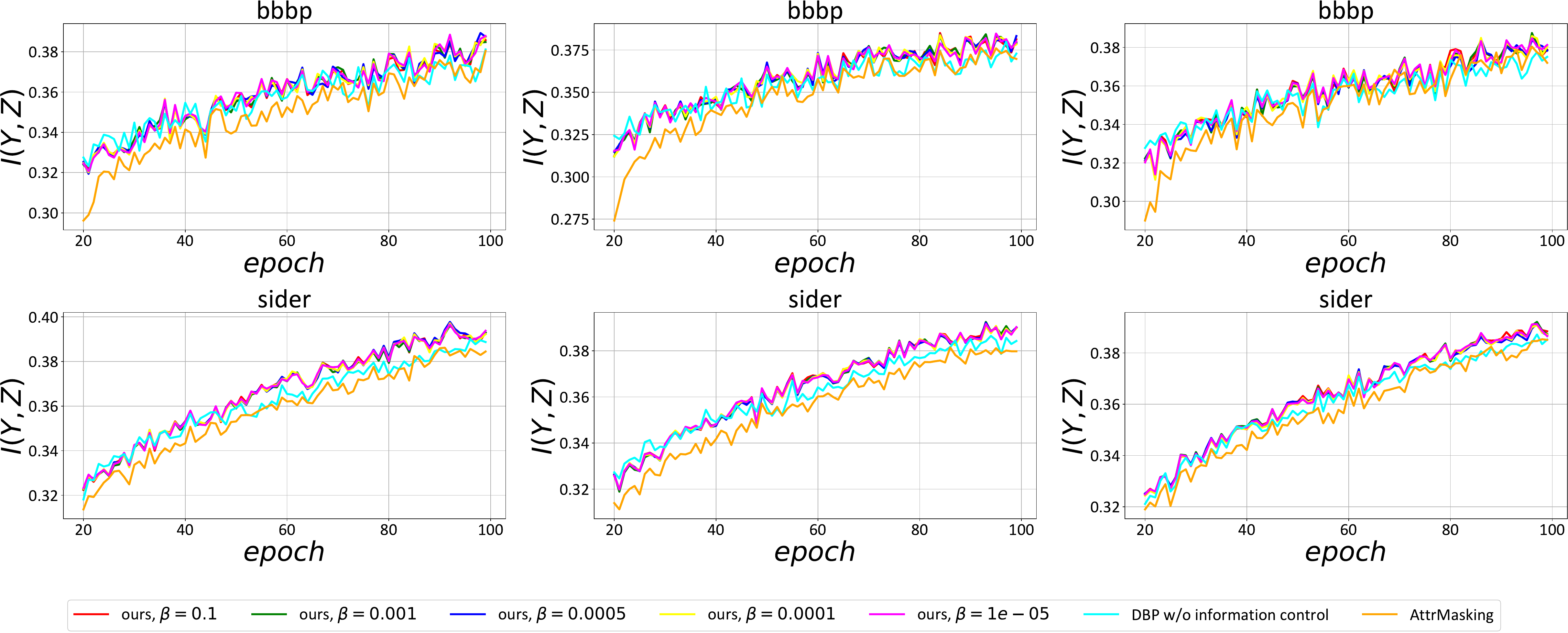}
    \caption{
    Dynamics of the mutual information $I(Y, Z)$ between the target labels $Y$ and the learned representations $Z$ across training epochs for different variants on two molecular property prediction datasets (BBBP and SIDER).
    }
    \label{fig:MI}
\end{figure*}
\begin{figure*}[!t]
    \centering  
    \includegraphics[width=1\linewidth]{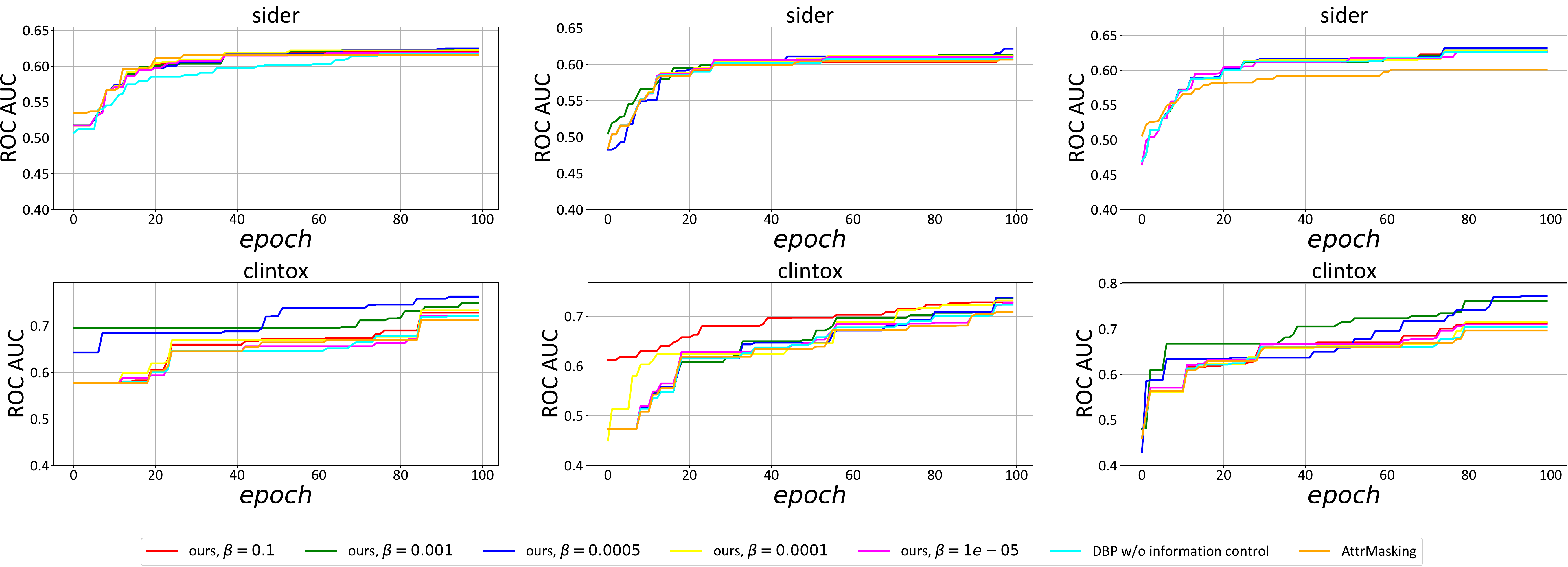}
    \caption{
    ROC-AUC curves across training epochs for different variants of the proposed DBP method on SIDER and ClinTox. 
    }
    \label{fig:ROC}
\end{figure*}

\subsection{Performance Comparison}
\noindent\textbf{Results on Chemistry Domain.} The performance comparison between the proposed DBP and SOTA methods on chemistry domain is shown in Table \ref{tbl:MPP}. DBP achieves the highest average ROC-AUC score and gain among all self-supervised learning strategies and performs best on six of eight tasks. We believe that such significant improvements can be attributed to the proposed information compression delayed pre-training strategy, which preserves more beneficial information in the pre-training phase and benefits downstream tasks in the fine-tuning phase. 

\noindent\textbf{Results on Biology Domain.} The results in Table \ref{tbl:BFP} show that DBP also achieves the best performance on biology domain, especially achieving a gain of 6.3\% compared to the No-pretrain baseline. This illustrates that the proposed strategy is general and generalizable. We argue that these properties are mainly affected by reasonable information control, which can learn more transferable prior knowledge and transfer them to the fine-tuning phase, thus benefiting more downstream tasks involving fine-grained classification.

\noindent\textbf{Results w.r.t. Different GNN Architectures.} Table \ref{tbl:GNN} compares the performance of DBP and state-of-the-art pre-training baselines, w.r.t. four different GNN architectures: GCN~\cite{DBLP:conf/iclr/KipfW17}, GraphSAGE~\cite{hamilton2017inductive}, GAT~\cite{velivckovic2018graph}, and GIN~\cite{DBLP:conf/iclr/XuHLJ19}. It can be observed that the proposed DBP consistently yields the best performance among all methods across architectures. This demonstrates that our strategy is pluggable and applicable to various GNN architectures. We deem that this improvement over previous works is mainly from the information compression delayed pre-training strategy in DBP, which is not included in existing methods.


\subsection{Analysis of the Training Process}
To gain a deeper understanding of our model, we analyzed various aspects of the training dynamics.

\noindent\textbf{Dynamics of Predictive Information.}
Figure \ref{fig:MI} illustrates the changes in mutual information between the representation $Z$ and the label $Y$ during the fine-tuning phase for different methods. Different values of $\beta$ represent the varying intensities of the information control during the fine-tuning phase, where the blue line represents the DBP without information control, and AttrMasking represents the traditional masking pre-training strategy. Across different variants (e.g., GCN, GAT, and GIN) and datasets (e.g., BBBP, ToxCast, SIDER, and ClinTox), these comparisons exhibit similar trends. Overall, the DBP method with information control strategies achieves higher mutual information $I(Y;Z)$ during the fine-tuning phase as compared to the traditional methods and DBP with less information control, which demonstrates the universality and effectiveness of our information control strategy across different datasets and models in enhancing the extraction of predictive relevant information during the fine-tuning stage.

\noindent\textbf{Dynamics of Performance Change.}
Figure \ref{fig:ROC} shows the changes in performance during the fine-tuning phase for different methods, using AUC-ROC as the performance metric. Similar to Figure~\ref{fig:MI}, different values of $\beta$ represent the varying intensities of information control during the fine-tuning phase, where the blue line represents the DBP without information control, and AttrMasking represents the traditional masking pre-training strategy. Across different variants and datasets, our strategy improves performance, but the performance ceiling is related to the choice of the intensity of information control $\beta$. For instance, on the ClinTox dataset, GCN exhibits the highest performance at $\beta=0.0005$, while GIN performs better at $\beta=0.1$. We will continue to analyze the sensitivity to $\beta$ in Section \ref{Other Analysis}.

\begin{figure*}[!t]
    \centering  
    \includegraphics[width=1\linewidth]{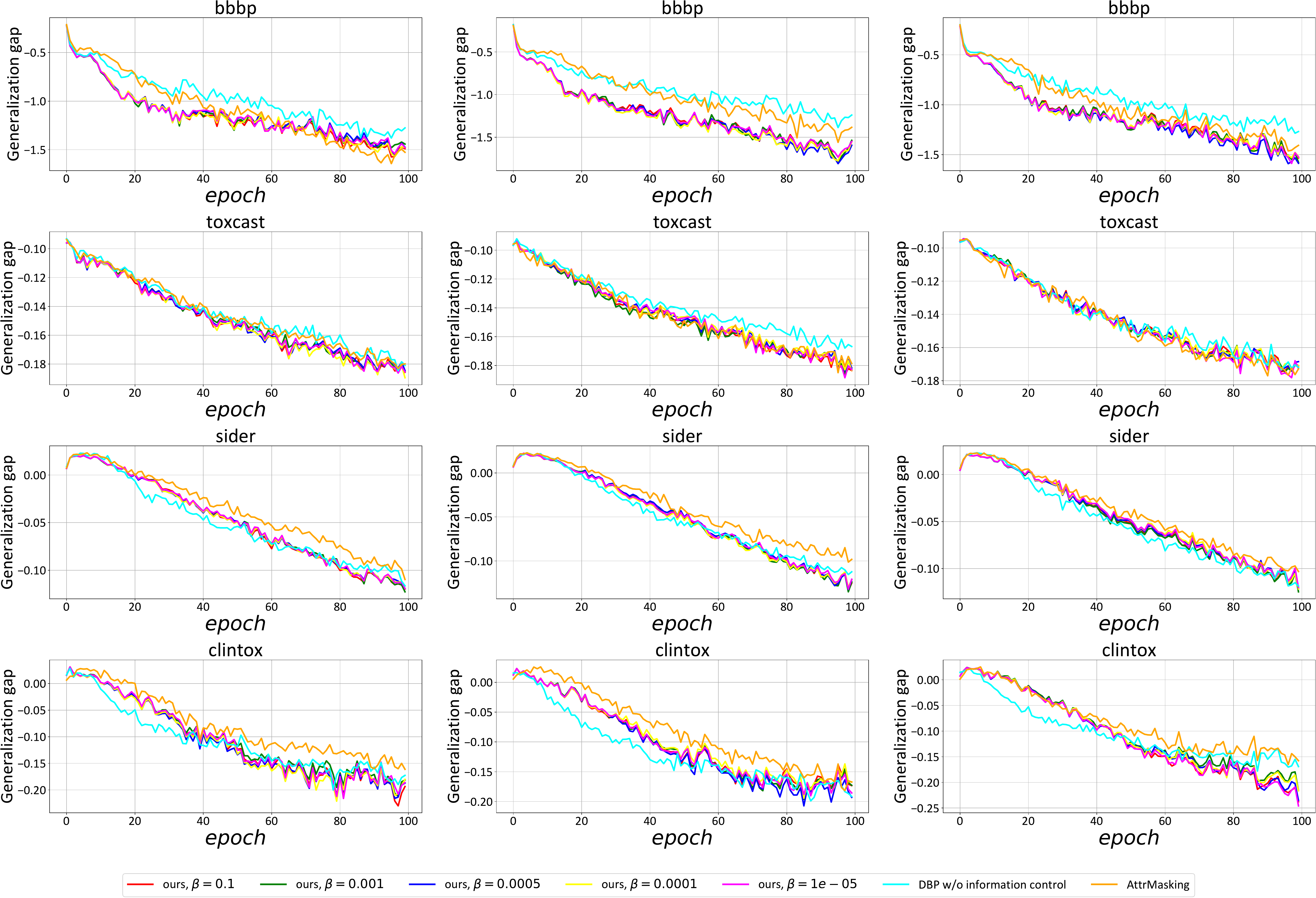}
    \caption{
    Generalization gap across training epochs for different variants of the proposed DBP method on four molecular property prediction datasets (BBBP, ToxCast, SIDER, and ClinTox).
    }
    \label{fig:Generalization}
\end{figure*}

\noindent\textbf{Dynamics of Generalization Gap.}
Figure \ref{fig:Generalization} demonstrates the changes in the generalization gap during the fine-tuning process for different methods. Similar to Figure \ref{fig:ROC}, different values of $\beta$ represent the varying intensities of the information control during the fine-tuning phase, where the blue line represents the DBP without information control, and AttrMasking represents the traditional masking pre-training strategy. Tests were conducted across different variants and datasets. Overall, our information control strategy achieves a lower generalization error during the fine-tuning phase compared to traditional methods and DBP with less information control. This relates to the conclusions drawn from Figure \ref{fig:MI}, as our information control strategy transfers more usable predictive relevant information, leading to the model learning more useful representations and thereby resulting in a smaller generalization error.
\begin{figure*}[!t]
    \centering  
    \includegraphics[width=1\linewidth]{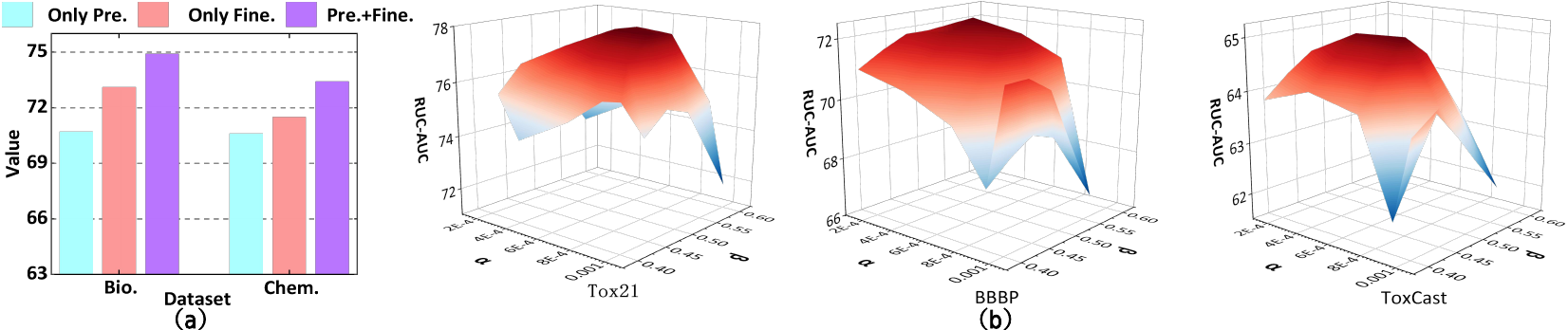}
    \caption{Hyperparameter sensitivity analysis and ablation study with respect to DBP. Subfigure (a) shows our ablation experiments on the information control modules during the pre-training and fine-tuning stages. Subfigure (b) illustrates our analysis experiments on the relationship between the information control hyperparameters $\alpha$ and $\beta$ and model performance across three datasets during the pre-training and fine-tuning stages.}
    \label{fig:Hyperparameter}
\end{figure*}

\begin{table}[!t]
	\begin{center}
    \caption{ROC-AUC scores (\%) on downstream biological function prediction task compared with state-of-the-art methods. \textbf{Bold} indicates the best performance while \underline{underline} indicates the second best.}
    \label{tbl:BFP}
	\scalebox{0.85}{
		\begin{tabular}{c| c c | c | c c}
			\toprule[2pt]
			\textbf{Method}  & \textbf{ROC-AUC} & \textbf{GAIN} & \textbf{Method}  & \textbf{ROC-AUC} & \textbf{GAIN} \\
			\midrule
            \midrule
			No Pre-training & 64.8 ± 1.0 & -   & EdgePred       & 65.9 ± 0.7 & 1.1 \\
			InfoGraph   & 65.1 ± 0.5 & 0.3 & AttrMasking    & 65.7 ± 0.5 & 0.9 \\
			ContextPred & 65.8 ± 0.3 & 1.0 & GraphPartition & 68.7 ± 0.2 & 3.9\\
            GraphCL     & 68.9 ± 0.6 & 4.1 & GraphLoG       & \underline{69.1 ± 0.7} & \underline{4.3}\\
            \midrule
            \rowcolor{gray!20}\multicolumn{2}{c|}{ \textbf{DBP} } & \multicolumn{2}{c}{\textbf{ROC-AUC:} \textbf{71.2 ± 0.4}} & \multicolumn{2}{c}{\textbf{GAIN:} \textbf{6.4}}  \\
			\bottomrule[2pt]
		\end{tabular}
	}
	\end{center}
\end{table}

\begin{table}[!t]
	\begin{center}
    \caption{ROC-AUC scores (\%) under various GNN architectures. All results are reported on biology domain. \textbf{Bold} indicates the best performance while \underline{underline} indicates the second best.}
    \label{tbl:GNN}
	\scalebox{1}{
		\begin{tabular}{c| c c c c}
			\toprule[2pt]
			\textbf{Method} & \textbf{GCN} & \textbf{GraphSAGE}& \textbf{GAT}& \textbf{GIN}\\
			\midrule
            \midrule
			EdgePred     & 64.7 ± 1.0 & 67.4 ± 1.5 & 67.4 ± 1.3 & 65.9 ± 1.7 \\
	      AttrMasking  & 64.4 ± 1.2 & 64.3 ± 0.8 & \underline{67.7 ± 1.2} & 65.7 ± 1.3 \\
	      ContextPred  & 64.6 ± 1.4 & 66.3 ± 0.7 & 66.9 ± 2.0 & 66.0 ± 1.2 \\
            GraphCL      & 67.6 ± 1.3 & 68.6 ± 0.4 & 67.2 ± 0.7 & 67.9 ± 0.9 \\
	      GraphLoG     & \underline{68.2 ± 0.6} & \underline{68.8 ± 0.8} & 67.5 ± 1.0 & \underline{69.1 ± 0.7} \\
            \midrule
            \rowcolor{gray!20} \textbf{DBP}    & \textbf{70.5 ± 0.7} & \textbf{70.2 ± 0.9} & \textbf{68.9 ± 1.3} & \textbf{71.2 ± 0.4}\\
			\bottomrule[2pt]
		\end{tabular}
	}
	\end{center}
\end{table}

\subsection{Further Analysis}\label{Other Analysis}
\noindent\textbf{Effect of Information Control Objective Function.} We attempt to analyze the effect of individually applying the information control objective function in the pre-training or fine-tuning phase on chemistry domain, with the same experimental setups as in Section \ref{sec:es}. As shown in Figure \ref{fig:Hyperparameter}(a), we have an interesting observation: When the pre-trained information control objective function is individually applied, its performance improvement on downstream tasks is limited, but applying the information control objective function independently in the fine-tuning phase can further improve the model performance. The reason might be that the information control objective in the fine-tuning phase is more conducive to improving classification performance, while the additional information retained by the information control module in the pre-training phase needs to be further compressed and selected before being applied to downstream tasks. 

\noindent\textbf{Analysis on Information Control Hyperparameters.} The hyperparameters $\alpha$ and $\beta$ that adjust the strength of information control are crucial to  DBP, so we further investigate the impact of different hyperparameter combinations on model performance. Experimental results on three downstream chemistry datasets are shown in Figure \ref{fig:Hyperparameter}(b). Interestingly, we observed that the performance of the model on downstream tasks degrades when the hyperparameters are too large or too small. We argue that excessively suppressing information compression in the pre-training phase will interfere with general knowledge extraction, while excessively enhancing information compression in the fine-tuning phase will make the latent representation less informative. 

\noindent\textbf{Comparison with Existing Work.}
 As graph pre-training research advances, some studies have noticed the negative impact of differences between pre-training and fine-tuning tasks, although they do not deeply analyze from an information and neural network training behavior perspective. For example, L2P-GNN\cite{lu2021learning} approaches from a meta-learning angle, simulating the fine-tuning process during pre-training to quickly adapt to new downstream tasks. GPPT \cite{sun2022gppt}using ideas from natural language processing's prompt learning, proposes a pre-training method focused on node classification, transforming downstream node classification tasks into edge prediction tasks similar to pre-training goals, bridging the gap between pre-training and fine-tuning objectives.
 
However, these methods primarily aim to reduce target differences during the fine-tuning stage, without analyzing the entire information transfer process across both stages. In contrast, our method analyzes the fundamental knowledge transfer from pre-training datasets to fine-tune downstream tasks from an information compression perspective. This framework could also be combined with downstream fine-tuning process variants like prompt learning and meta-learning in the future.

\noindent\textbf{Complexity Analysis.}
The DBP framework encompasses pre-training and fine-tuning phases. Pre-training involves self-supervised and reconstruction tasks, both with a computational complexity of $O(V+E)$. Fine-tuning introduces an information compression module and a classifier, each with a complexity of $O(V)$. Overall, DBP maintains a complexity of $O(V+E)$, similar to traditional GNNs.

\section{Conclusions}
 In this paper, we reexamine the pre-training process within the traditional pre-training and fine-tuning framework from the perspective of Information Bottleneck, and confirm that the forgetting phenomenon in the pre-training phase exactly has detrimental effects on downstream tasks. Then, we propose a DBP framework that maintains as much as possible mutual information during the pre-training phase by suppressing the compression operation and delays the compression operation to fine-tuning phase. To achieve this, we design two information control objectives that can be directly optimized and further integrate them into the model design. Extensive experiments demonstrate the effectiveness and generalization of DBP.

\section{Acknowledgment}
This research was supported by the National Natural Science Foundation of China (Grants No. 62072427, No. 12227901), the Project of Stable Support for Youth Team in Basic Research Field, CAS (No. YSBR-005), and the Academic Leaders Cultivation Program, USTC. Additionally, we acknowledge funding from the Research Grants Council of the Hong Kong Special Administrative Region, China [GRF Project No. Cityu11215723], and the Key Basic Research Foundation of Shenzhen, China (JCYJ20220818100005011).

\bibliographystyle{IEEEtran}
\bibliography{ijcal_new}

\begin{thebibliography}{10}
\providecommand{\url}[1]{#1}
\csname url@samestyle\endcsname
\providecommand{\newblock}{\relax}
\providecommand{\bibinfo}[2]{#2}
\providecommand{\BIBentrySTDinterwordspacing}{\spaceskip=0pt\relax}
\providecommand{\BIBentryALTinterwordstretchfactor}{4}
\providecommand{\BIBentryALTinterwordspacing}{\spaceskip=\fontdimen2\font plus
\BIBentryALTinterwordstretchfactor\fontdimen3\font minus \fontdimen4\font\relax}
\providecommand{\BIBforeignlanguage}[2]{{%
\expandafter\ifx\csname l@#1\endcsname\relax
\typeout{** WARNING: IEEEtran.bst: No hyphenation pattern has been}%
\typeout{** loaded for the language `#1'. Using the pattern for}%
\typeout{** the default language instead.}%
\else
\language=\csname l@#1\endcsname
\fi
#2}}
\providecommand{\BIBdecl}{\relax}
\BIBdecl

\bibitem{bhagat2011node}
S.~Bhagat, G.~Cormode, and S.~Muthukrishnan, ``Node classification in social networks,'' in \emph{Social network data analytics}, 2011.

\bibitem{he2020bi}
J.~He, H.~Liu, Y.~Zheng, S.~Tang, W.~He, and X.~Du, ``Bi-labeled lda: Inferring interest tags for non-famous users in social network,'' \emph{Data Science and Engineering}, 2020.

\bibitem{wang2021tedic}
Y.~Wang, P.~Li, C.~Bai, and J.~Leskovec, ``Tedic: Neural modeling of behavioral patterns in dynamic social interaction networks,'' in \emph{Proc. of WWW}, 2021.

\bibitem{zhang2016understanding}
F.~Zhang, J.~Zhai, B.~He, S.~Zhang, and W.~Chen, ``Understanding co-running behaviors on integrated cpu/gpu architectures,'' \emph{IEEE Transactions on Parallel and Distributed Systems}, 2016.

\bibitem{zhang2020gcn}
S.~Zhang, H.~Yin, T.~Chen, Q.~V.~N. Hung, Z.~Huang, and L.~Cui, ``Gcn-based user representation learning for unifying robust recommendation and fraudster detection,'' in \emph{Proc. of SIGIR}, 2020.

\bibitem{gilmer2017neural}
J.~Gilmer, S.~S. Schoenholz, P.~F. Riley, O.~Vinyals, and G.~E. Dahl, ``Neural message passing for quantum chemistry,'' in \emph{Proc. of ICML}, 2017.

\bibitem{liu2019n}
S.~Liu, M.~F. Demirel, and Y.~Liang, ``N-gram graph: Simple unsupervised representation for graphs, with applications to molecules,'' \emph{Proc. of NeurIPS}, 2019.

\bibitem{wu2023chemistry}
Z.~Wu, J.~Wang, H.~Du, D.~Jiang, Y.~Kang, D.~Li, P.~Pan, Y.~Deng, D.~Cao, C.-Y. Hsieh \emph{et~al.}, ``Chemistry-intuitive explanation of graph neural networks for molecular property prediction with substructure masking,'' \emph{Nature Communications}, vol.~14, no.~1, p. 2585, 2023.

\bibitem{reiser2022graph}
P.~Reiser, M.~Neubert, A.~Eberhard, L.~Torresi, C.~Zhou, C.~Shao, H.~Metni, C.~van Hoesel, H.~Schopmans, T.~Sommer \emph{et~al.}, ``Graph neural networks for materials science and chemistry,'' \emph{Communications Materials}, vol.~3, no.~1, p.~93, 2022.

\bibitem{li2022graph}
Z.~Li, K.~Meidani, P.~Yadav, and A.~Barati~Farimani, ``Graph neural networks accelerated molecular dynamics,'' \emph{The Journal of Chemical Physics}, vol. 156, no.~14, p. 144103, 2022.

\bibitem{wang2019knowledge}
H.~Wang, F.~Zhang, M.~Zhang, J.~Leskovec, M.~Zhao, W.~Li, and Z.~Wang, ``Knowledge-aware graph neural networks with label smoothness regularization for recommender systems,'' in \emph{Proc. of KDD}, 2019.

\bibitem{ying2018graph}
R.~Ying, R.~He, K.~Chen, P.~Eksombatchai, W.~L. Hamilton, and J.~Leskovec, ``Graph convolutional neural networks for web-scale recommender systems,'' in \emph{Proc. of KDD}, 2018.

\bibitem{yang2023dgrec}
L.~Yang, S.~Wang, Y.~Tao, J.~Sun, X.~Liu, P.~S. Yu, and T.~Wang, ``Dgrec: Graph neural network for recommendation with diversified embedding generation,'' in \emph{Proceedings of the Sixteenth ACM International Conference on Web Search and Data Mining}, 2023, pp. 661--669.

\bibitem{chang2021sequential}
J.~Chang, C.~Gao, Y.~Zheng, Y.~Hui, Y.~Niu, Y.~Song, D.~Jin, and Y.~Li, ``Sequential recommendation with graph neural networks,'' in \emph{Proceedings of the 44th international ACM SIGIR conference on research and development in information retrieval}, 2021, pp. 378--387.

\bibitem{hao2023multi}
Y.~Hao, J.~Ma, P.~Zhao, G.~Liu, X.~Xian, L.~Zhao, and V.~S. Sheng, ``Multi-dimensional graph neural network for sequential recommendation,'' \emph{Pattern Recognition}, vol. 139, p. 109504, 2023.

\bibitem{gao2022graph}
C.~Gao, X.~Wang, X.~He, and Y.~Li, ``Graph neural networks for recommender system,'' in \emph{Proceedings of the Fifteenth ACM International Conference on Web Search and Data Mining}, 2022, pp. 1623--1625.

\bibitem{shi2020graph}
C.~Shi, M.~Xu, H.~Guo, M.~Zhang, and J.~Tang, ``A graph to graphs framework for retrosynthesis prediction,'' in \emph{Proc. of ICML}, 2020.

\bibitem{shi2019graphaf}
C.~Shi, M.~Xu, Z.~Zhu, W.~Zhang, M.~Zhang, and J.~Tang, ``Graphaf: a flow-based autoregressive model for molecular graph generation,'' in \emph{Proc. of ICLR}, 2019.

\bibitem{you2018graph}
J.~You, B.~Liu, Z.~Ying, V.~Pande, and J.~Leskovec, ``Graph convolutional policy network for goal-directed molecular graph generation,'' \emph{Proc. of NeurIPS}, 2018.

\bibitem{ji2021survey}
S.~Ji, S.~Pan, E.~Cambria, P.~Marttinen, and S.~Y. Philip, ``A survey on knowledge graphs: Representation, acquisition, and applications,'' \emph{IEEE Transactions on Neural Networks and Learning Systems}, 2021.

\bibitem{wang2023joint}
P.~Wang, C.~Ge, Z.~Zhou, X.~Wang, Y.~Li, and Y.~Wang, ``Joint gated co-attention based multi-modal networks for subregion house price prediction,'' \emph{IEEE Transactions on Knowledge \& Data Engineering}, vol.~35, no.~02, pp. 1667--1680, 2023.

\bibitem{zhao2024twist}
Z.~Zhao, P.~Wang, H.~Wen, Y.~Zhang, Z.~Zhou, and Y.~Wang, ``A twist for graph classification: Optimizing causal information flow in graph neural networks,'' in \emph{Proceedings of the AAAI Conference on Artificial Intelligence}, vol.~38, no.~15, 2024, pp. 17\,042--17\,050.

\bibitem{hu2020gpt}
Z.~Hu, Y.~Dong, K.~Wang, K.-W. Chang, and Y.~Sun, ``Gpt-gnn: Generative pre-training of graph neural networks,'' in \emph{Proc. of KDD}, 2020.

\bibitem{DBLP:conf/kdd/HouLCDYW022}
Z.~Hou, X.~Liu, Y.~Cen, Y.~Dong, H.~Yang, C.~Wang, and J.~Tang, ``Graphmae: Self-supervised masked graph autoencoders,'' in \emph{Proc. of KDD}, 2022.

\bibitem{hu2020strategies}
W.~Hu, B.~Liu, J.~Gomes, M.~Zitnik, P.~Liang, V.~Pande, and J.~Leskovec, ``Strategies for pre-training graph neural networks,'' in \emph{Proc. of ICLR}, 2020.

\bibitem{you2020graph}
Y.~You, T.~Chen, Y.~Sui, T.~Chen, Z.~Wang, and Y.~Shen, ``Graph contrastive learning with augmentations,'' \emph{Advances in neural information processing systems}, vol.~33, pp. 5812--5823, 2020.

\bibitem{xu2021self}
M.~Xu, H.~Wang, B.~Ni, H.~Guo, and J.~Tang, ``Self-supervised graph-level representation learning with local and global structure,'' in \emph{Proc. of ICML}, 2021.

\bibitem{sun2020infograph}
F.-Y. Sun, J.~Hoffman, V.~Verma, and J.~Tang, ``Infograph: Unsupervised and semi-supervised graph-level representation learning via mutual information maximization,'' in \emph{Proc. of ICLR}, 2020.

\bibitem{you2021graph}
Y.~You, T.~Chen, Y.~Shen, and Z.~Wang, ``Graph contrastive learning automated,'' in \emph{Proc. of ICML}, 2021.

\bibitem{wu2021self}
L.~Wu, H.~Lin, C.~Tan, Z.~Gao, and S.~Z. Li, ``Self-supervised learning on graphs: Contrastive, generative, or predictive,'' \emph{IEEE Transactions on Knowledge and Data Engineering}, 2021.

\bibitem{tran2022s5cl}
M.~Tran, S.~J. Wagner, M.~Boxberg, and T.~Peng, ``S5cl: Unifying fully-supervised, self-supervised, and semi-supervised learning through hierarchical contrastive learning,'' in \emph{Medical Image Computing and Computer Assisted Intervention--MICCAI 2022: 25th International Conference, Singapore, September 18--22, 2022, Proceedings, Part II}.\hskip 1em plus 0.5em minus 0.4em\relax Springer, 2022, pp. 99--108.

\bibitem{zhu2021graph}
Y.~Zhu, Y.~Xu, F.~Yu, Q.~Liu, S.~Wu, and L.~Wang, ``Graph contrastive learning with adaptive augmentation,'' in \emph{Proceedings of the Web Conference 2021}, 2021, pp. 2069--2080.

\bibitem{verma2021graphmix}
V.~Verma, M.~Qu, K.~Kawaguchi, A.~Lamb, Y.~Bengio, J.~Kannala, and J.~Tang, ``Graphmix: Improved training of gnns for semi-supervised learning,'' in \emph{Proceedings of the AAAI conference on artificial intelligence}, vol.~35, no.~11, 2021, pp. 10\,024--10\,032.

\bibitem{liu2023graphprompt}
Z.~Liu, X.~Yu, Y.~Fang, and X.~Zhang, ``Graphprompt: Unifying pre-training and downstream tasks for graph neural networks,'' in \emph{Proceedings of the ACM Web Conference 2023}, 2023, pp. 417--428.

\bibitem{li2021targeted}
C.~Li and Z.~Qiu, ``Targeted bert pre-training and fine-tuning approach for entity relation extraction,'' in \emph{Data Science: 7th International Conference of Pioneering Computer Scientists, Engineers and Educators, ICPCSEE 2021, Taiyuan, China, September 17--20, 2021, Proceedings, Part II 7}.\hskip 1em plus 0.5em minus 0.4em\relax Springer, 2021, pp. 116--125.

\bibitem{wan2022factpegasus}
D.~Wan and M.~Bansal, ``Factpegasus: Factuality-aware pre-training and fine-tuning for abstractive summarization,'' \emph{arXiv preprint arXiv:2205.07830}, 2022.

\bibitem{anderson2022prefrontal}
M.~C. Anderson and S.~B. Floresco, ``Prefrontal-hippocampal interactions supporting the extinction of emotional memories: the retrieval stopping model,'' \emph{Neuropsychopharmacology}, vol.~47, no.~1, pp. 180--195, 2022.

\bibitem{gruber2019cultural}
T.~Gruber, L.~Luncz, J.~M{\"o}rchen, C.~Schuppli, R.~L. Kendal, and K.~Hockings, ``Cultural change in animals: a flexible behavioural adaptation to human disturbance,'' \emph{Palgrave Communications}, vol.~5, no.~1, 2019.

\bibitem{kitazono2017multiple}
T.~Kitazono, S.~Hara-Kuge, O.~Matsuda, A.~Inoue, M.~Fujiwara, and T.~Ishihara, ``Multiple signaling pathways coordinately regulate forgetting of olfactory adaptation through control of sensory responses in caenorhabditis elegans,'' \emph{Journal of Neuroscience}, vol.~37, no.~42, pp. 10\,240--10\,251, 2017.

\bibitem{gravitz2019importance}
L.~Gravitz, ``The importance of forgetting,'' \emph{Nature}, 2019.

\bibitem{achille2018critical}
A.~Achille, M.~Rovere, and S.~Soatto, ``Critical learning periods in deep networks,'' in \emph{Proc. of ICLR}, 2018.

\bibitem{shwartz2017opening}
R.~Shwartz-Ziv and N.~Tishby, ``Opening the black box of deep neural networks via information (2017),'' \emph{arXiv preprint arXiv:1703.00810}, 2017.

\bibitem{li2020enhanced}
J.~Li, H.~Xu, S.-Y. Sun, S.~Liu, N.~Li, Q.~Li, H.~Liu, and Z.~Li, ``Enhanced spiking neural network with forgetting phenomenon based on electronic synaptic devices,'' \emph{Neurocomputing}, vol. 408, pp. 21--30, 2020.

\bibitem{peng2021learning}
J.~Peng, X.~Sun, M.~Deng, C.~Tao, B.~Tang, W.~Li, G.~Wu, Y.~Liu, T.~Lin, H.~Li \emph{et~al.}, ``Learning by active forgetting for neural networks,'' \emph{arXiv preprint arXiv:2111.10831}, 2021.

\bibitem{cossu2022continual}
A.~Cossu, T.~Tuytelaars, A.~Carta, L.~Passaro, V.~Lomonaco, and D.~Bacciu, ``Continual pre-training mitigates forgetting in language and vision,'' \emph{arXiv preprint arXiv:2205.09357}, 2022.

\bibitem{feng2020codebert}
Z.~Feng, D.~Guo, D.~Tang, N.~Duan, X.~Feng, M.~Gong, L.~Shou, B.~Qin, T.~Liu, D.~Jiang \emph{et~al.}, ``Codebert: A pre-trained model for programming and natural languages,'' \emph{arXiv preprint arXiv:2002.08155}, 2020.

\bibitem{velickovic2019deep}
P.~Velickovic, W.~Fedus, W.~L. Hamilton, P.~Li{\`o}, Y.~Bengio, and R.~D. Hjelm, ``Deep graph infomax.'' \emph{ICLR (Poster)}, 2019.

\bibitem{liu2021pre}
S.~Liu, H.~Wang, W.~Liu, J.~Lasenby, H.~Guo, and J.~Tang, ``Pre-training molecular graph representation with 3d geometry,'' \emph{arXiv preprint arXiv:2110.07728}, 2021.

\bibitem{DBLP:conf/iclr/AlemiFD017}
A.~A. Alemi, I.~Fischer, J.~V. Dillon, and K.~Murphy, ``Deep variational information bottleneck,'' in \emph{Proc. of ICLR}, 2017.

\bibitem{wu2020graph}
T.~Wu, H.~Ren, P.~Li, and J.~Leskovec, ``Graph information bottleneck,'' \emph{Proc. of NeurIPS}, 2020.

\bibitem{yu2020graph}
J.~Yu, T.~Xu, Y.~Rong, Y.~Bian, J.~Huang, and R.~He, ``Graph information bottleneck for subgraph recognition,'' in \emph{Proc. of ICLR}, 2020.

\bibitem{peng2020graph}
Z.~Peng, W.~Huang, M.~Luo, Q.~Zheng, Y.~Rong, T.~Xu, and J.~Huang, ``Graph representation learning via graphical mutual information maximization,'' in \emph{Proc. of WWW}, 2020.

\bibitem{sterling2015zinc}
T.~Sterling and J.~J. Irwin, ``Zinc 15--ligand discovery for everyone,'' \emph{Journal of chemical information and modeling}, 2015.

\bibitem{wu2018moleculenet}
Z.~Wu, B.~Ramsundar, E.~N. Feinberg, J.~Gomes, C.~Geniesse, A.~S. Pappu, K.~Leswing, and V.~Pande, ``Moleculenet: a benchmark for molecular machine learning,'' \emph{Chemical science}, 2018.

\bibitem{kingma2015adam}
D.~P. Kingma and J.~Ba, ``Adam: A method for stochastic optimization,'' in \emph{ICLR (Poster)}, 2015.

\bibitem{DBLP:conf/iclr/KipfW17}
T.~N. Kipf and M.~Welling, ``Semi-supervised classification with graph convolutional networks,'' in \emph{Proc. of ICLR}, 2017.

\bibitem{hamilton2017inductive}
W.~Hamilton, Z.~Ying, and J.~Leskovec, ``Inductive representation learning on large graphs,'' \emph{Proc. of NeurIPS}, 2017.

\bibitem{velivckovic2018graph}
P.~Veli{\v{c}}kovi{\'c}, G.~Cucurull, A.~Casanova, A.~Romero, P.~Li{\`o}, and Y.~Bengio, ``Graph attention networks,'' in \emph{Proc. of ICLR}, 2018.

\bibitem{DBLP:conf/iclr/XuHLJ19}
K.~Xu, W.~Hu, J.~Leskovec, and S.~Jegelka, ``How powerful are graph neural networks?'' in \emph{Proc. of ICLR}, 2019.

\bibitem{liu2022unsupervised}
N.~Liu, S.~Jian, D.~Li, and H.~Xu, ``Unsupervised hierarchical graph pooling via substructure-sensitive mutual information maximization,'' in \emph{Proc. of CIKM}, 2022.

\bibitem{you2020does}
Y.~You, T.~Chen, Z.~Wang, and Y.~Shen, ``When does self-supervision help graph convolutional networks?'' in \emph{Proc. of ICML}, 2020.

\bibitem{hou2022graphmae}
Z.~Hou, X.~Liu, Y.~Cen, Y.~Dong, H.~Yang, C.~Wang, and J.~Tang, ``Graphmae: Self-supervised masked graph autoencoders,'' in \emph{Proceedings of the 28th ACM SIGKDD Conference on Knowledge Discovery and Data Mining}, 2022, pp. 594--604.

\bibitem{tan2023s2gae}
Q.~Tan, N.~Liu, X.~Huang, S.-H. Choi, L.~Li, R.~Chen, and X.~Hu, ``S2gae: self-supervised graph autoencoders are generalizable learners with graph masking,'' in \emph{Proceedings of the sixteenth ACM international conference on web search and data mining}, 2023, pp. 787--795.

\bibitem{lu2021learning}
Y.~Lu, X.~Jiang, Y.~Fang, and C.~Shi, ``Learning to pre-train graph neural networks,'' in \emph{Proceedings of the AAAI conference on artificial intelligence}, vol.~35, no.~5, 2021, pp. 4276--4284.

\bibitem{sun2022gppt}
M.~Sun, K.~Zhou, X.~He, Y.~Wang, and X.~Wang, ``Gppt: Graph pre-training and prompt tuning to generalize graph neural networks,'' in \emph{Proceedings of the 28th ACM SIGKDD Conference on Knowledge Discovery and Data Mining}, 2022, pp. 1717--1727.

\end{thebibliography}
\vspace{-0.35in}
\begin{IEEEbiography}[{\includegraphics[width=1in,height=1.25in]{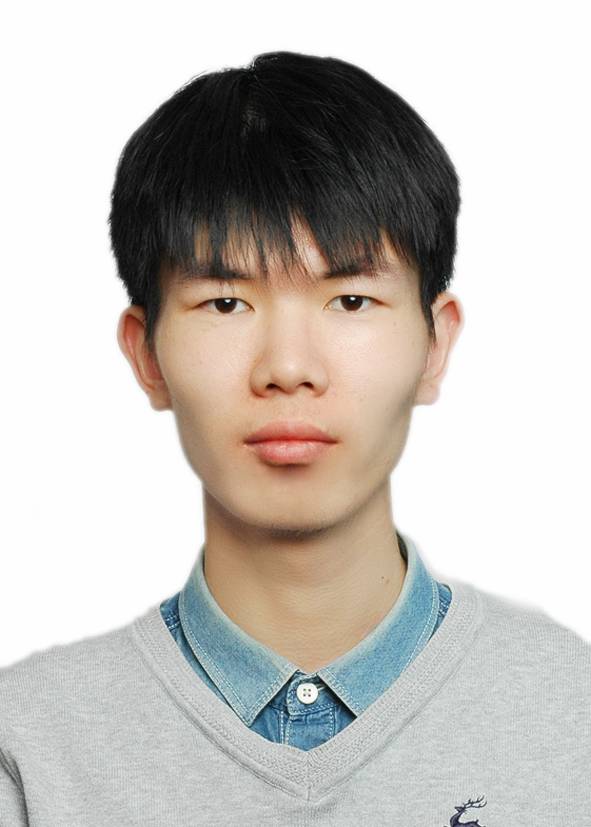}}]{Zhe Zhao} is now a doctoral student in the School of Data Science, University of Science and Technology of China and computer science school, City University of Hong Kong. He received his bachelor degree of computer science from Anhui University in 2021. His research interests mainly include data mining, machine learning, and multi-objective optimization.
\end{IEEEbiography}
\vspace{-0.15in}
\begin{IEEEbiography}[{ \centering \includegraphics[width=1in,height=1.35in]{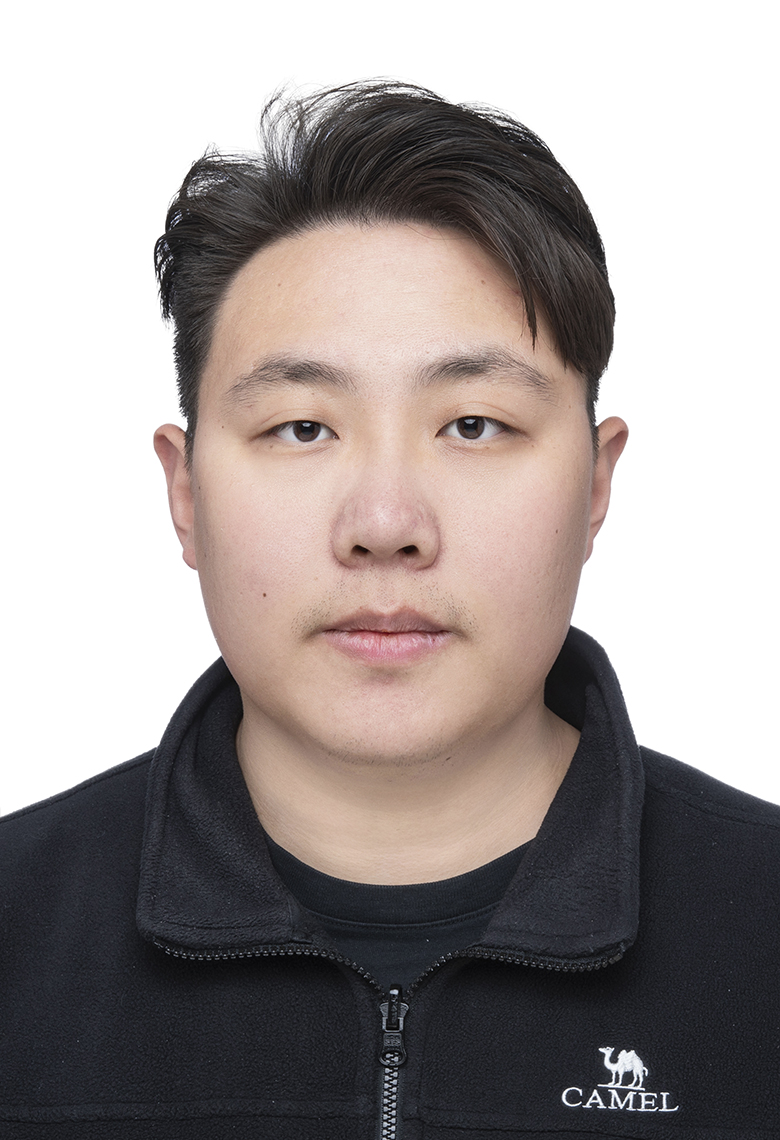}}]{Pengkun Wang}(Member, IEEE)
	is now an associate researcher at Suzhou Institute for Advanced Research, University of Science and Technology of China (USTC). He got his Ph.D. degree at University of Science and Technology of China in 2023. He has published over 30 papers on top conferences and journals such as IEEE TKDE, AAAI, ICLR, KDD, and WWW. His mainly research interests include imbalanced/long-tailed learning, interpretable data augmentation, multi-objective optimization, and open environment data mining. He also works on applying generalized deep learning to real-world tasks such as urban computing and AI4Science.
\end{IEEEbiography}
\vspace{-0.15in}
\begin{IEEEbiography}[{\includegraphics[width=1in,height=1.25in]{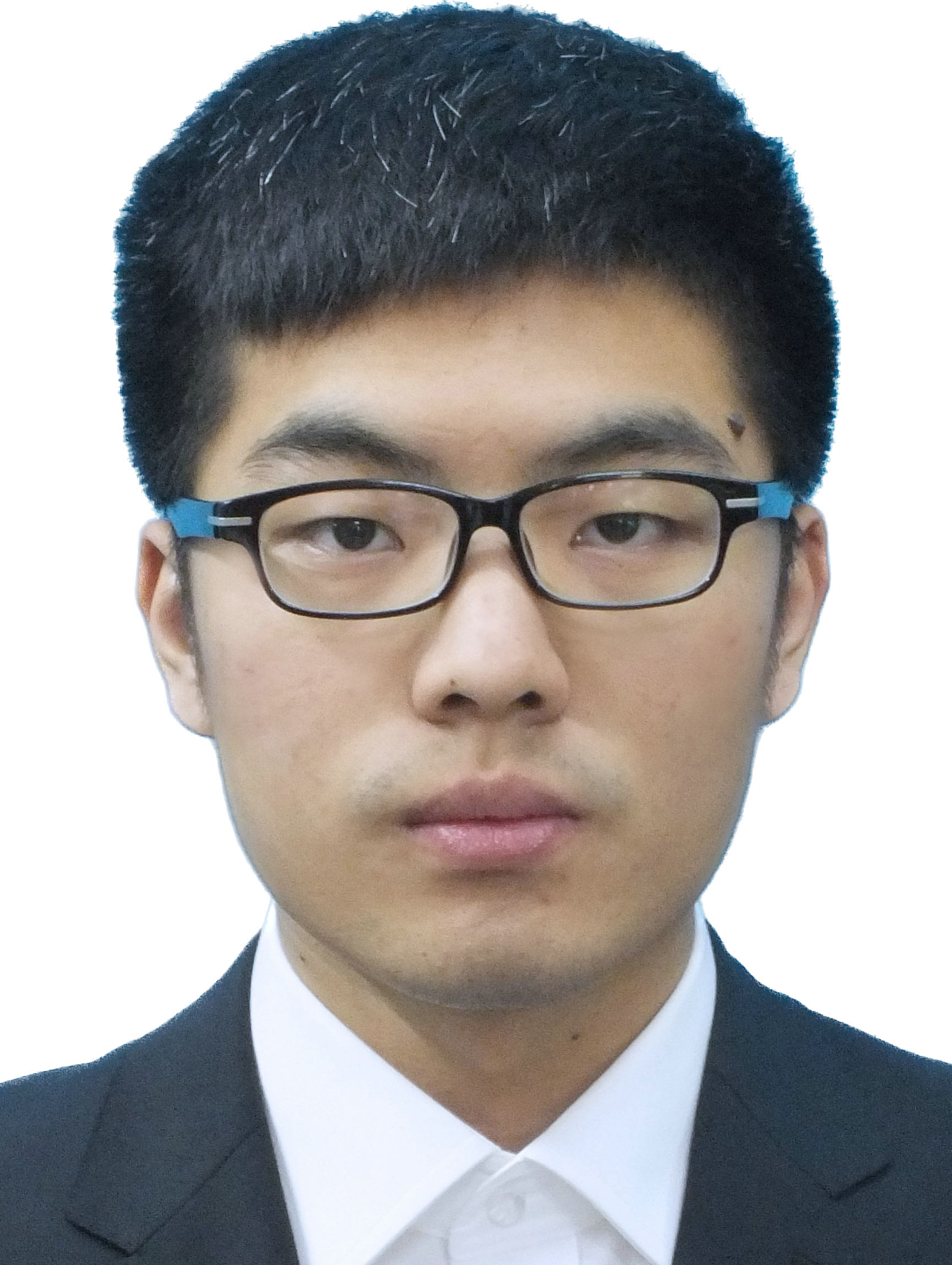}}]{Xu Wang}
	is now an associate researcher at Suzhou Institute for Advanced Research, University of Science and Technology of China (USTC). He got his Ph.D. degree at University of Science and Technology of China in 2023. He has published over 10 papers on top conferences and journals such as KDD, IEEE TVT, and WSDM. His mainly research interests include spatiotemporal data mining, time series analysis and AI for science.
\end{IEEEbiography}
\vspace{-0.1in}
\begin{IEEEbiography}[{\includegraphics[width=1in,height=1.25in]{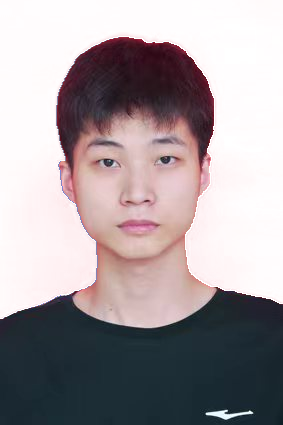}}]{Haibin Wen} is currently pursuing his undergraduate studies at Shaoguan University, with research interests encompassing high-performance computing and neural networks.
\end{IEEEbiography}
\vspace{-0.15in}
\begin{IEEEbiography}[{\includegraphics[width=1in,height=1.25in]{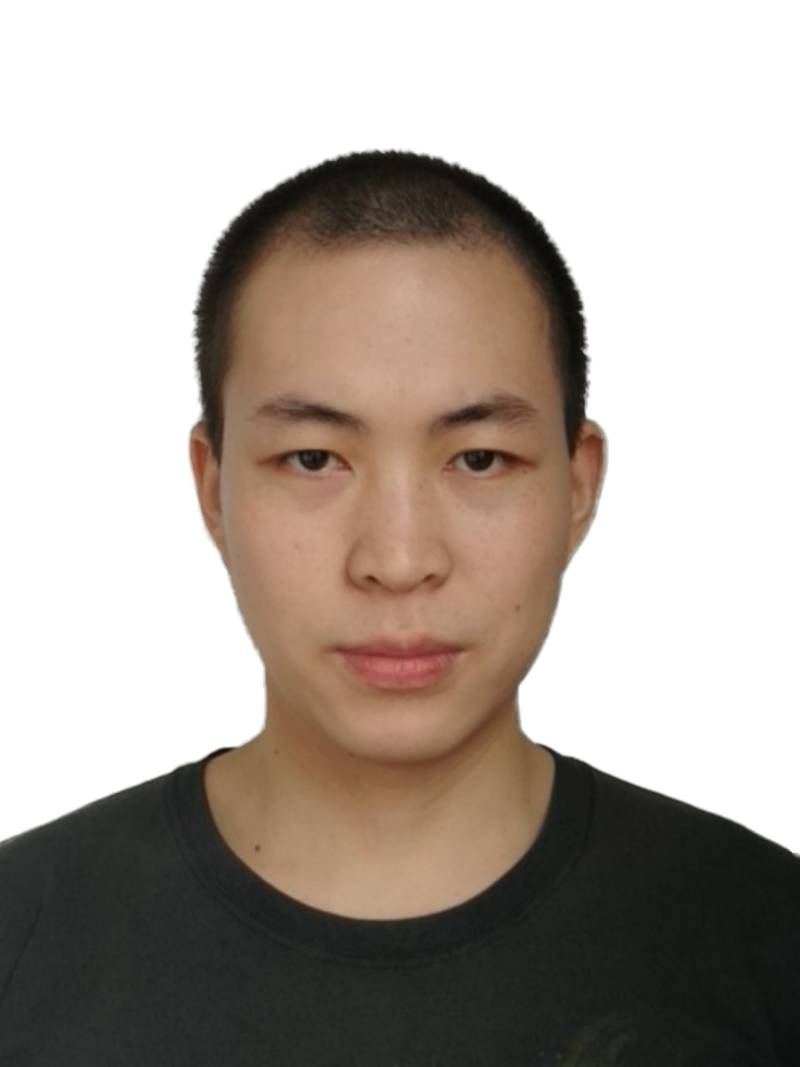}}]{Xiaolong Xie} graduated from Nanchang University in 2021 with an undergraduate degree and is currently pursuing a graduate degree in Software Engineering there. He works in the Emergency Rescue VR Laboratory, focusing on neural radiation fields and 3D reconstruction. His contributions to various projects have earned him accolades from supervisors and peers.
\end{IEEEbiography}
\vspace{-0.1in}
\begin{IEEEbiography}[{\includegraphics[width=1in,height=1.25in]{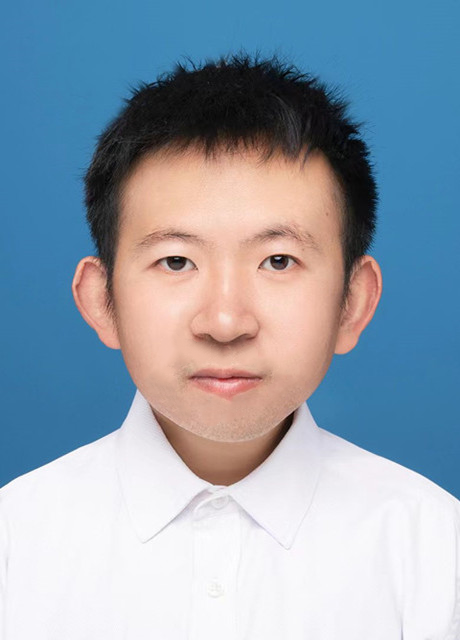}}]{Zhengyang Zhou}(Member, IEEE) is now an associate researcher at Suzhou Institute for Advanced Research, University of Science and Technology of China (USTC). He got his Ph.D. degree at University of Science and Technology of China in 2023. He has published over 30 papers on top conferences and journals such as NeurIPS, ICLR, KDD, TKDE, WWW, AAAI, and ICDE. His mainly research interests include spatiotemporal data mining, human-centered urban computing, and deep learning generalization with model behavior analysis. He is now especially interested in improving generalization capacity of neural networks for streaming and spatiotemporal data.
\end{IEEEbiography}
\vspace{-0.1in}
\begin{IEEEbiography}[{\includegraphics[width=1in,height=1.25in]{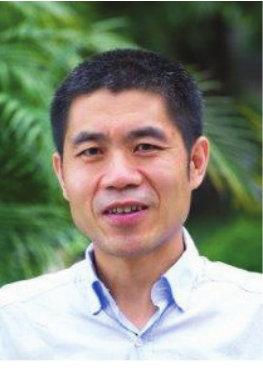}}]{Qingfu Zhang}(Fellow, IEEE) received the B.S. degree in mathematics from Shanxi University, Taiyuan, China, in 1984, and the M.S. degree in applied mathematics and the Ph.D. degree in information engineering from Xidian University, Xi’an, China, in 1991 and 1994, respectively. He is a Chair Professor of Computational Intelligence with the Department of Computer
Science, City University of Hong Kong, Hong Kong. His main research interests include evolutionary computation, optimization, metaheuristics, and their applications.
Dr. Zhang was awarded the 2010 IEEE \textsc{Transactions on Evolutionary Computation} Outstanding Paper Award. He is an Associate Editor of the IEEE \textsc{Transactions on Evolutionary Computation} and the IEEE \textsc{Transactions on Cyberntics}.
\end{IEEEbiography}
\vspace{-0.1in}
\begin{IEEEbiography}[{\includegraphics[width=1in,height=1.25in]{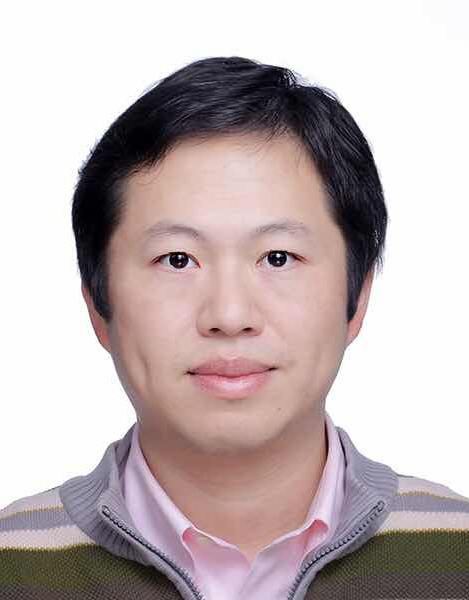}}]{Yang Wang}(Senior Member, IEEE)
	is now an associate professor at School of Computer Science and Technology, School of Software Engineering, and School of Data Science in USTC. He got his Ph.D. degree at University of Science and Technology of China in 2007. Since then, he keeps working at USTC till now as a postdoc and an associate professor successively. Meanwhile, he also serves as the vice dean of school of software engineering of USTC. His research interest mainly includes wireless (sensor) networks, distribute systems, data mining, and machine learning, and he is also interested in all kinds of applications of AI and data mining technologies, especially in urban computing and AI4Science.
\end{IEEEbiography}

\vfill

\end{document}